\documentclass[aoas]{imsart}

\RequirePackage{amsthm,amsmath,amsfonts,amssymb}
\RequirePackage[authoryear]{natbib}
\RequirePackage{graphicx}
\usepackage{subfigure}
\usepackage{pifont} 
\usepackage{float}
\startlocaldefs

\endlocaldefs

\begin{document}

\begin{frontmatter}
\title{$l_{1-2}$ GLasso: $L_{1-2}$ Regularized Multi-task Graphical Lasso for Joint Estimation of eQTL Mapping and Gene Network}
\runtitle{The Graphical Lasso Based on Difference of $l_{1}$ and $l_{2}$ Norms}

\begin{aug}
\author[A]{\fnms{Wei}~\snm{Miao}\ead[label=e1]{miaow@hnu.edu.cn}},
\author[B]{\fnms{Lan}~\snm{Yao}\ead[label=e2]{yao@hnu.edu.cn}\orcid{0000-0000-0000-0000}}
\address[A]{College of Mathematics,
Hunan University\printead[presep={,\ }]{e1}}

\address[B]{College of Mathematics,
Hunan University\printead[presep={,\ }]{e2}}
\end{aug}

\begin{abstract}
A critical problem in genetics is to discover how gene expression is regulated within cells. 
Two major tasks of regulatory association learning are : (i) identifying SNP-gene relationships, known as eQTL mapping, and (ii) determining gene-gene relationships, known as gene network estimation. 
To share information between these two tasks, we focus on the unified model for joint estimation of eQTL mapping and gene network, and propose a $L_{1-2}$ regularized multi-task graphical lasso, named $L_{1-2}$ GLasso.
Numerical experiments on artificial datasets demonstrate the competitive performance of $L_{1-2}$ GLasso on capturing the true sparse structure of eQTL mapping and gene network. $L_{1-2}$ GLasso is further applied to real dataset of ADNI-1 and experimental results show that $L_{1 -2}$ GLasso can obtain sparser and more accurate solutions than other commonly-used methods.

\end{abstract}


\end{frontmatter}

\section{Introduction}
	
    Developments in sequencing technology allow us to obtain more and more genomic data since the publication of the first human genome sequence. 
	Computational techniques can help us to mine meaningful information from raw data and understand how gene expression is regulated in cells.
	In general, these problems include identifying cancer gene co-expression (co-expression: simultaneous expression of two or more genes) modules, determining SNP-gene relationships through eQTL (expression quantitative trait locus) mapping and determining gene-gene relationships by estimating gene network structure, etc \citep{rockman2006genetics,gardner2005reverse}.  
	Given a dataset containing single nucleotide polymorphisms (SNPs) and mRNA expression, the problem is to understand the SNP-gene and gene-gene relationships. 
	For example, assuming SNPs $x=(x_{1},\dots,x_{p})$ and genes $y=(y_{1},\dots,y_{q})$, the SNP-gene relationships in eQTL mapping are determined by a regression coefficient matrix and the gene-gene relationships in gene network estimation are captured by output structure. 

	There have been many types of research on eQTL mapping and gene network estimation. The traditional method of eQTL mapping is to determine whether there is an association between a gene and an SNP. Later, multivariate models have been developed to determine relationships between multiple SNPs and a gene \citep{michaelson2010data}. More recently, several models have been proposed to determine relationships between multiple SNPs and multiple genes \citep{kim2012tree}.  
	
	As for gene network estimation, the traditional method is to construct a graph and connect two related genes with an edge. To be specific, many previous studies inferred gene-gene relationships from gene expression data. For example, in Gaussian Graphical Models (GGM) framework, graphical models use graphs to represent dependencies between random variables \citep{schafer2005shrinkage, segal2005signatures, li2006gradient, peng2009partial}. In GGM, multivariate vectors follow a multivariate normal distribution and have a specific structure of the covariance matrix. The inverse of the covariance matrix is called the concentration matrix. GGM assumes that the expression variation pattern of a given gene can be predicted by a small subset of other genes \citep{meinshausen2006high}. The assumption leads to the sparsity (i.e., multiple zeros) in the concentration matrix and reduces the problem to a well-known neighborhood selection or covariance selection problem. In the concentration map modeling framework, the key idea is to use a partial correlation as a measure of the independence of any two genes, thereby directly distinguishing between direct and indirect interactions. In other approaches, Bayesian Networks are also utilized to establish the structure between genes \citep{marbach2010revealing}.
    
    The Multi-task regression model can be used to jointly estimate the regression coefficient matrix and the output structure. One challenge to be faced is the high-dimensional data disaster which is very common in genetic data. In previous studies, sparse learning is a good way to deal with this problem and has attracted wide attention due to its advantages of sparse solutions, strong interpretability, and convenient computation \citep{bertsimas2020sparse}. 
    Furthermore, to enhance the expression ability, researchers have proposed various structured sparse models which combine sparse learning with structured regularization. 
	In various fields of computing and engineering, it is an important research topic to construct a structured sparse model based on the prior assumption of sparsity and the specific structural characteristics of the problem. 
	
	Many models which are based on structured sparsity regularization have been reviewed \citep{vinga2021structured}. Group Lasso, which encourages related exit groups to have nonzero coefficients for the same subset of inputs, has been studied extensively \citep{yuan2006model}. A computationally efficient way was provided to perform Lasso-regularized estimation of sparse concentration matrices \citep{friedman2008sparse}. Graph-Guide Fused Lasso encourages pairs of outputs linked in a graph to have similar coefficient values \citep{kim2009statistical}. Conditional Gaussian Graphical Models (CGGM) have been developed to estimate both the output structure and the regression coefficients with structured sparsity at the same time \citep{yin2011sparse,li2012sparse,chun2013joint}. However, these models all require us to have prior knowledge of relationships between the output $y$.
    Another class of models, which focuses on estimating the conditional covariance of $y|x$ rather than the covariance structure of the output $y$, has been developed to learn both the regression coefficient matrix and the output structure \citep{rothman2010sparse}. Under the influence of noisy data, these models may not end up with the true structure between outputs.
    Recently, a novel approach called Inverse-Covariance-Fused Lasso (ICLasso) which focuses on the covariance structure of the output $y$, can also jointly estimate the regression coefficient matrix and the output structure \citep{marchetti2019penalized}. The structured sparsity regularization penalty is formed by the $l_{1}$ norm in ICLasso.
    
	In addition, many other regularization penalties have also been studied. The projection operators that can enforce both $l_{1}$ and $l_{2}$ norms have been developed for encouraging sparsity in structured sparse models \citep{hoyer2004non}. One of the regularization penalties that has been studied a lot is the difference of $l_{1}$ and $l_{2}$ norms. The penalty is considered robust and can help select sparse solutions \citep{yin2014ratio}. It has been used in nonnegative least squares (NNLS) and orthogonal matching pursuit. The comparisons with $l_{1}$ minimization for imaging data can be found in \citep{esser2013method}. Some researchers have also applied the difference of $l_{1}$ and $l_{2}$ norms in sparse signal reconstruction problems to approximate the original $l_{0}$-norm-based sparseness \citep{liu2016and}. It can be seen in other areas such as compressed sensing, seismic inversions, etc \citep{yin2015minimization,wang2019three}.
	
	Motivated by these studies, we propose a new model based on difference of $l_{1}$ and $l_{2}$ norms. Our model makes some important new contributions: 
	(\romannumeral1) We introduce a new regularization penalty into the model inducing a better approximation and use a faster algorithm to solve the optimization problem.
	(\romannumeral2) Under the same parameter setting, the solved regression coefficient matrix is sparser compared with existing methods such as MRCE, ICLasso, etc.  
	(\romannumeral3) Our model outperforms other baseline methods in the recovery of the output structure.
	
	In Section 2, we give an introduction to several baseline methods. In Section 3, we describe our new model with a different penalty and the optimization algorithm in detail. In sections 4 and 5, we evaluate the effectiveness of our method on simulated and real datasets. Finally, we summarize the article in section 6.
	
	\section{Background}
	We assume that $X \in R^{n \times p}$ is a matrix of SNP genotypes and $Y \in R^{n \times q}$ is a matrix of gene expression values. Here, $n$ represents the number of samples, $q$ represents the number of genes, and $p$ represents the number of SNPs. We show the exact matrix form as:
	\begin{equation}\label{ma}
	\underbrace{\left[
        \begin{array}{cccc}
           y_{11} & y_{12} & \cdots & y_{1q} \\
	       y_{21} & y_{22} & \cdots & y_{2q} \\
	       \vdots&\vdots& \ddots&\vdots \\
	       y_{n1} & y_{n2} & \cdots & y_{nq} \\
         \end{array}
     \right]}_{Y}
     =
     \underbrace{\left[
     \begin{array}{cccc}
           x_{11} & x_{12} & \cdots & x_{1p} \\
	       x_{21} & x_{22} & \cdots & x_{2p} \\
	       \vdots&\vdots& \ddots&\vdots \\
	       x_{n1} & x_{n2} & \cdots & x_{np} \\
         \end{array}
     \right]}_{X}
     \times
     \underbrace{\left[
     \begin{array}{cccc}
           \beta_{11} & \beta_{12} & \cdots & \beta_{1q} \\
	       \beta_{21} & \beta_{22} & \cdots & \beta_{2q} \\
	       \vdots&\vdots& \ddots&\vdots \\
	       \beta_{p1} & \beta_{p2} & \cdots & \beta_{pq} \\
         \end{array}
     \right]}_{B}
\end{equation}

	\subsection{Multi-Task Lasso}
	Multi-Task Lasso can be used for statistical tests to detect SNPs that are associated with genes \citep{tibshirani1996regression}. Given $X$ and $Y$, the multivariate linear regression model is given by  
	
	\begin{equation}
		y_{k}=X \beta_{k}+\epsilon_{k}, k=1, \ldots, q,
	\end{equation}where $\beta_{k}=[\beta_{1k},\dots,\beta_{pk}]^T$ represents regression coefficients. It can be used to detect SNPs that are significantly associated with genes. We assume $\epsilon_{k}\sim{N(0,\sigma^{2})}$ and the mathematical expression in matrix form is:
	
	\begin{equation}\label{M1}
		\min_{B}~ \frac{1}{n}\|Y-X B\|_{F}^{2}+\lambda\|B\|_{1},
	\end{equation}where $B \in R^{p \times q}$ represents the regression coefficient matrix, $\lambda$ is the regularization parameter, which is used to control the degree of sparsity. 

	\subsection{Multivariate regression with covariance estimation}
	Multivariate regression with covariance estimation (MRCE) is a method that can jointly estimate the regression coefficient matrix and the output structure \citep{rothman2010sparse}.
    It assumes that $X$ has the linear relationship with $Y$: $Y = XB + E$, in which $E \sim \mathcal{N}\left(0, \Omega^{-1}\right)$ is a Gaussian noise matrix. We can calculate that $Y \mid X \sim \mathcal{N}\left(X B, \Omega^{-1}\right)$. MRCE can be expressed as follows:
	 
	 \begin{equation}
     \begin{aligned}
      \min _{B, \Omega}~ & \frac{1}{n} \operatorname{tr}\left((Y-XB)^{T}(Y-XB) \Omega\right) \\
      &-\log \operatorname{det}(\Omega)+\lambda_{1}\|B\|_{1}+\lambda_{2}\|\Omega\|_{1},
     \end{aligned}
     \label{MRCE}
     \end{equation}
     where $\Omega$ represents the conditional inverse covariance of $Y|X$ rather than the inverse covariance of $Y$.
     Based on previous assumptions, $\Omega$ is related to the Gaussian noise matrix $E$ and it can not capture the exact relationship between the regression coefficient matrix $B$ and the output structure in $Y$.

	\subsection{Inverse-Covariance-Fused Lasso}
	
	\cite{marchetti2019penalized} introduced a new model called Inverse-Covariance-Fused Lasso. The model can also jointly estimate regression coefficients and the output structure. Compared with previous studies, the method captures the marginal inverse covariance of $Y$ rather than the conditional inverse covariance of $Y|X$. 
	
	ICLasso (Inverse-Covariance-Fused Lasso) begins with two core modeling assumptions: 
	\begin{center}
		$x \sim \mathcal{N}(0,T)$\\
		$y|x \sim \mathcal{N}(x^{T}B,E)$,
	\end{center}where $T$ represents the covariance of $x$ and $E$ represents the conditional covariance of $y|x$, $\Theta$ represents the marginal inverse covariance of $Y$, which is different from $\Omega$ in (\ref{MRCE}). 
	
	With these assumptions, we can derive the marginal distribution of $y$. Based on the fact that the marginal distribution $p(y)$ follows the Gaussian distribution, then use the law of total expectation and the law of total variance to derive the mean and covariance of $y$, as follows:
	
	\begin{equation}\label{e6}
		\begin{aligned}
			\mathbb{E}_{y}(y)=\mathbb{E}_{x}\left(\mathbb{E}_{y \mid x}(y \mid x)\right)=\mathbb{E}_{x}\left(x^{T}B)\right)=0\\
			\operatorname{Cov}_{y}(y)=\mathbb{E}_{x}\left(\operatorname{Cov}_{y \mid x}(y \mid x)\right)+\operatorname{Cov}_{x}\left(\mathbb{E}_{y \mid x}(y \mid x)\right)\\
			=\mathbb{E}_{x}\left(E\right)+\operatorname{Cov}_{x}\left(x^{T}B\right)=E+B^{T} T B.
		\end{aligned}
	\end{equation}
	
	We can calculate the distribution of $y$ :
	
	\begin{equation}
		y \sim \mathcal{N}\left(0, \Theta^{-1}\right),
	\end{equation}where $\Theta^{-1}=E+B^{T} T B$ is the marginal covariance of $y$. This is a connection between the inverse covariance of $y$ and the regression coefficient matrix $B$. 
	For simplicity, we assume $T=\tau^2 I_{p\times p}$ and $E=\varepsilon^2 I_{q\times q}$ and then $\Theta^{-1} \propto B^{T} B$.
	
	Given i.i.d. observations of SNPs $x \in R^{p}$ and genes $y \in R^{q}$, in order to jointly estimate the regression coefficient matrix $B \in R^{p \times q}$ and the inverse covariance matrix $\Theta \in R^{q \times q}$,
	the inverse-covariance-fused lasso optimization problem can be written as:
	
	\begin{equation}
		\begin{aligned}
			\min_{B, \Theta}~ & \frac{1}{n}\|Y-X B\|_{F}^{2}+\frac{1}{n} \operatorname{tr}\left(Y^{T} Y \Theta\right)-\log \operatorname{det}(\Theta) \\
			&+\lambda_{1}\|B\|_{1}+\lambda_{2}\|\Theta\|_{1} \\
			&+\gamma \sum_{(k, m)}\left|\theta_{k m}\right| \cdot\left\|\beta_{.k}+\operatorname{sgn}\left(\theta_{k m}\right) \beta_{.m}\right\|_{1}.
		\end{aligned}
	\end{equation}
	
	This objective effectively boils down to a combination of problems: Multi-Task Lasso and Sparse Inverse Covariance Estimation including a graph-guided fusion penalty form. The role of the penalty $\sum_{(k, m)}\left|\theta_{k m}\right| \cdot\left\|\beta_{.k}+\operatorname{sgn}\left(\theta_{k m}\right) \beta_{.m}\right\|_{1}$ is to encourage structural information sharing between $B$ and $\Theta$.
	
	\subsection{Estimating Model Parameters with a Fusion Penalty}
	
	Previous studies provide us with an idea to estimate these parameters and also encourage information sharing between $B$ and $\Theta$. To do this, the model is formulated as a convex optimization problem, whose objective function is:
	
	\begin{equation} \label{penalty}
		\operatorname{loss}_{y|x}(B)+\operatorname{loss}_{y}(\Theta)+\operatorname{penalty}(B,\Theta),
	\end{equation}where we can see: 
	
	$\bullet$  $\operatorname{loss}_{y|x}(B)$ can be derived from the negative log-likelihood of $y|x$;
	   
	$\bullet$  $\operatorname{loss}_{y}(\Theta)$ can be derived from the negative marginal log-likelihood of $y$;
	
	$\bullet$  $\operatorname{penalty}(B,\Theta)$ is a penalty term that encourages shared information between the regression coefficient matrix $B$ and the output structure $\Theta$.
	
	The $l_{1}$ norm penalty $\|B\|_{1}$ and $\|\Theta\|_{1}$ induce sparsity in the estimates of $B$ and $\Theta$, which make the model feasible even on high dimensional data. To obtain a sparser solution, we can naturally generalize the $l_{1}$ norm penalty. 
	Based on this framework, we propose our model in the next section.
	
	\section{The $l_{1-2}$ Graphical Lasso}
	
    Since the sparsity of $B$ is reflected by the number of its nonzero terms, it is equivalent to the so-called $l_{0}$ norm penalty. ICLasso replaces $l_{0}$ norm penalty with $l_{1}$ norm penalty. A reconstruction framework based on the difference $l_ {1}$ and $l_{2}$ norms was proposed \citep{esser2013method}. It can be reformulated as follows:
 	
 	\begin{equation}
 		\begin{array}{ll}
 			\min_{x} & \lambda\|x\|_{1}-\tau\|x\|_{2} \\
 			\text { s.t. } & y = Ax.
 		\end{array}
 	\end{equation}
    
    It was proved that as long as the selection of appropriate $\lambda$ and $\tau$, the solution can be close to the solution of the problem with $l_{0}$ norm penalty.
    To get sparser solutions, we propose the $l_{1-2}$ Graphical Lasso. Our model can be described as:

	\begin{equation} \label{l1-l2penalty}
		\operatorname{loss}_{y|x}(B)+\operatorname{loss}_{y}(\Theta)+\operatorname{penalty}(B,\Theta)+\operatorname{penalty}(B)+\operatorname{penalty}(\Theta).
	\end{equation}
	Specifically, the expression is as followed:
	   \begin{equation} \label{l12}
		\begin{aligned}
			\min_{B, \Theta}~ & \frac{1}{n}\|Y-X B\|_{F}^{2}+\frac{1}{n} \operatorname{tr}\left(Y^{T} Y \Theta\right)-\log \operatorname{det}(\Theta) \\
			&+\lambda_{1}||B||_{1}-\tau||B||_{2,1}+\lambda_{2}\|\Theta\|_{1} \\
			&+\gamma \sum_{(k, m)}\left|\theta_{k m}\right| \cdot\left\|\beta_{. k}+\operatorname{sgn}\left(\theta_{k m}\right) \beta_{.m}\right\|_{1}.
		\end{aligned} 
	\end{equation}
 
    We define:
	\begin{equation}\label{L1-L21}
		g_{12}(B)=\frac{1}{n}||Y-XB||_{F}^{2}+\lambda_{1}||B||_{1}-\tau||B||_{2,1}	
	\end{equation}
	
	\begin{equation}\label{L1-L22}
		h(\Theta)=\frac{1}{n}tr(Y^{T}Y\Theta)-\log \operatorname{det}(\Theta)+\lambda_{2}||\Theta||_{1}	
	\end{equation}
    
    \begin{equation}\label{L1-L23}
        \operatorname{GFL}(B,-\Theta)=\sum_{k=1}^q \sum_{m=1}^q\left|\theta_{k m}\right| \cdot\left\|\beta_{\cdot k}+\operatorname{sgn}\left(\theta_{k m}\right) \beta_{\cdot m}\right\|_1
    \end{equation}
    The term $\frac{1}{n}\|Y-X B\|_{F}^{2}$, $\frac{1}{n} \operatorname{tr}\left(Y^{T} Y \Theta\right)-\log \operatorname{det}(\Theta)$  and $\gamma \sum_{(k, m)}\left|\theta_{k m}\right| \cdot\left\|\beta_{k}+\operatorname{sgn}\left(\theta_{k m}\right) \beta_{.m}\right\|_{1}$ are derived from Equation (\ref{penalty}) respectively.
    We describe the role of each item in detail: 
    
    $\bullet$  $\operatorname{loss}_{y|x}(B)$: $\frac{1}{n}\|Y-X B\|_{F}^{2}$. According to $y \mid x \sim \mathcal{N}\left(x^{T} B, \varepsilon^{2} I\right)$, we can derive its expression by Maximum Likelihood Estimation. The role of this term is to encourage the regression coefficient matrix $B$.
    
    $\bullet$  $\operatorname{loss}_{y}(\Theta)$: $\frac{1}{n} \operatorname{tr}\left(Y^{T} Y \Theta\right)-\log \operatorname{det}(\Theta)$. According to $y \sim \mathcal{N}\left(0, \Theta^{-1}\right)$, we can derive its expression by Maximum Likelihood Estimation. The role of this term is to encourage the inverse covariance $\Theta$ to reflect the correlations among the outputs.
    
    $\bullet$  $\operatorname{penalty}(B,\Theta)$: $\gamma \sum_{(k, m)}\left|\theta_{k m}\right| \cdot\left\|\beta_{\cdot k}+\operatorname{sgn}\left(\theta_{k m}\right) \beta_{\cdot m}\right\|_{1}$ is a graph-guided fusion penalty. It encourages the regression coefficients of closely related outputs to be similar. When $y_{k}$ is partially positively correlated with $y_{m}$ and $\beta_{jk} \neq \beta_{jm}$ for any $j$, it imposes a penalty proportional to $\theta_{km}$, and when $y_{k}$ and $y_{m}$ have a negative partial correlation for any $j$, it imposes a penalty proportional to $-\theta_{km}$.
    
    $\bullet$  $\operatorname{penalty}(B)$: $\lambda_{1}||B||_{1}-\tau||B||_{2,1}=\lambda_{1} \sum_{j, k}\left|\beta_{j k}\right|+\tau_{2} \sum_{j=1}^{p}\sqrt{\sum_{k=1}^{q}\beta_{j k}^{2}}$ is an $l_{1-2}$ norm penalty over the regression coefficient matrix that induces sparsity in $B$. Compared to $l_{1}$ norm, the difference of $l_{1}$ and $l_{2}$ norms is closer to the $l_{0}$ norm.  
      
    $\bullet$  $\operatorname{penalty}(\Theta)$: $\lambda_{2}\|\Theta\|_{1}=\lambda_{2} \sum_{k, m}\left|\theta_{k m}\right|$ is an $l_{1}$ norm penalty that induces sparsity in $\Theta$.
    
	\subsection{Relationship to ICLasso}
	$l_{1 \mbox{-} 2}$-GLasso also implicitly assumes two underlying modeling assumptions: $x \sim \mathcal{N}(0,T)$ and $y|x \sim \mathcal{N}(x^{T}B,E)$. The difference between $l_{1 \mbox{-} 2}$-GLasso and ICLasso is that we use difference of $l_{1}$ and $l_{2}$ norms to get sparse solutions that are closer to the real world. The experimental results are shown in the next section. We find that $l_{1 \mbox{-} 2}$-GLasso can obtain sparser solutions than other models while maintaining the regression error.  
	
	\subsection{Optimization}
	Previous work on the problem (\ref{l12}) propose some off-the-shelf algorithms to solve the above optimization problem. 
	Based on these studies, we use the alternating minimization strategy to solve the $l_{1 \mbox{-} 2}$-GLasso.
	Pay attention to the $\operatorname{GFL}(B,-\Theta)$ term, it is clear that this term is a bi-convex function. Thus, upon defining
	
	\begin{equation}{\label{joint}}
		\begin{aligned}
			&g_{12}(B)=\frac{1}{n}\|Y-XB\|_{F}^{2}+\lambda_{1}\|B\|_{1}-\tau||B||_{2,1} \\
			&h(\Theta)=\frac{1}{n} \operatorname{tr}\left(Y^{T} Y \Theta\right)-\log \operatorname{det}(\Theta)+\lambda_{2}\|\Theta\|_{1}.
		\end{aligned}
	\end{equation}
	
	The original objective can be rewritten as   
	
	\begin{equation}
		\min_{B, \Theta}~ g_{12}(B)+h(\Theta)+\operatorname{GFL}(B,-\Theta).
	\end{equation}
	
	Compared with ICLasso, $l_{1 \mbox{-} 2}$-GLasso can obtain sparser regression coefficient matrix by difference of $l_{1}$ and $l_{2}$ norms. The difficulty of solving the problem also lies in this penalty. It can be seen that the term $\operatorname{GFL}(B,-\Theta)$ is bi-convex, so we can use an alternating minimization strategy developed by \cite{marchetti2019penalized} to solve the problem (\ref{joint}). 
    
    First, we fix $\Theta$, so the problem becomes:
	
	\begin{equation}
		f_{\Theta}(B)=g_{12}(B)+\operatorname{GFL}(B,-\Theta).
	\end{equation}
	
	Although our model introduces a new regularization term $\lambda_{1}||B||_{1}-\tau||B||_{2,1}$, it can be decomposed into the form of a differentiable function and a non-differentiable function. For the matrix $B = [\boldsymbol{\beta_{1}}, \boldsymbol{\beta_{2}}, \dots, \boldsymbol{\beta_{p}}]^{T}$, $\boldsymbol{\beta_{i}}^{T}$ is the $ith$ row in $B$. According to the definition:
	\begin{equation}
		||B||_{2,1}=\sum_{i=1}^{p}(\boldsymbol{\beta_{i}}^{T}\boldsymbol{\beta_{i}})^{\frac{1}{2}},
	\end{equation}
    we define the $\Sigma$ and the derivative of $||B||_{2,1}$ as:
    \begin{equation}
    \Sigma:=
	\begin{bmatrix}
           \frac{1}{||\boldsymbol{\beta_{1}}||_{2}} & 0 & \cdots & 0 \\
	       0 & \frac{1}{||\boldsymbol{\beta_{2}}||_{2}} & \cdots & 0 \\
	       \vdots&\vdots& \ddots&\vdots \\
	       0 & 0 & \cdots & \frac{1}{||\boldsymbol{\beta_{p}}||_{2}} \\
         \end{bmatrix}, 
	\end{equation}
    \begin{equation}
		\frac{ \partial ||B||_{2,1} }{ \partial B } = (\frac{ \partial \sum_{i=1}^{p}(\boldsymbol{\beta_{i}}^{T}\boldsymbol{\beta_{i}})^{\frac{1}{2}} }{ \partial \boldsymbol{\beta_{i}} })_{p \times 1} = \begin{bmatrix}
           \frac{1}{||\boldsymbol{\beta_{1}}||_{2}} & 0 & \cdots & 0 \\
	       0 & \frac{1}{||\boldsymbol{\beta_{2}}||_{2}} & \cdots & 0 \\
	       \vdots&\vdots& \ddots&\vdots \\
	       0 & 0 & \cdots & \frac{1}{||\boldsymbol{\beta_{p}}||_{2}} \\
         \end{bmatrix} B = \Sigma B.
	\end{equation}
 
   When we use a small $\tau$, the term $\|Y-XB\|_{F}^{2}-\tau||B||_{2,1}$ is considered convex. This problem can be solved by the proximal-average proximal gradient descent (PA-PG) algorithm \cite{yu2013better}. Compared to the proximal gradient descent, PA-PG converges consistently faster. It was proved that with a suitable stepsize, the subgradient method converges in at most $O(1/\epsilon)$ steps for any accuracy $\epsilon>0$. First, the derivation of $\|Y-XB\|_{F}^{2}-\tau||B||_{2,1}$ is $ X^{T}(XB-Y)-\tau \Sigma$, then we take a gradient step of the form $B-\nu (X^{T}(XB-Y)-\tau \Sigma)$ and some small step size $\nu$ are used. Other optimization procedures can follow \cite{marchetti2019penalized}.
   Finally, this sub-problem can be written as 
	
	\begin{equation}
		\widetilde{\beta}_{jk},\widetilde{\beta}_{jm}=\rm{arg}\mathop{\min}\limits_{\beta_{jk},\beta_{jm}}\frac{1}{2\nu}(\beta_{jk}-z_{jk})^{2}+(\beta_{jm}-z_{jm})^{2}+|\beta_{jk}+sgn(\theta_{km})\beta_{jm}|.
	\end{equation}
	
	We can find a closed-form solution for $\beta$. The solution to this sub-problem is relatively simple, and the convergence speed of the algorithm can also be found in \cite{yu2013better}.

	Then fix $B$, and the problem becomes:
	
	\begin{equation} \label{subp2}
		\mathop{\min}\limits_{\Theta}~ h(\Theta)+\operatorname{GFL}(B,-\Theta).
	\end{equation}
	
	This problem can be solved by adapting the block coordinate descent (BCD) algorithm \citep{friedman2008sparse}. 
	Finally, this sub-problem can be written as:
	
 	\begin{equation}
 		\mathop{\min}\limits_{\alpha}~ \frac{1}{2}\alpha^{T}\widetilde{H}_{j}\alpha+u^{T}\alpha_{+}-l^{T}\alpha_{-}.
 	\end{equation}
	
    We solve each coordinate using the coordinate descent method and applying a variant of the soft threshold operator.

	\section{Simulation Study}
	In this section, we compare different models on synthetic data with known values of $B$ and $\Theta$, so that we can directly measure how well the true parameter values are recovered. Models include Graph-Guided Fused Lasso (GFLasso), Sparse Multivariate Regression with Covariance Estimation (MRCE), Inverse-Covariance-Fused Lasso (ICLasso), and $l_{1 \mbox{-} 2}$-GLasso. 
	For each model, we select hyperparameter values by minimizing the error on a held-out validation set. 
	
	We evaluate each model from two dimensions: (\romannumeral1) Recovery of sparse structures. (\romannumeral2) Regression error of $B$ and $\Theta$.
	
	(\romannumeral1) \textbf{Recovery of sparse structures}: To evaluate how well each model can estimate the sparsity structure of $B$ and $\Theta$, we calculate the F1 score for recovering the elements of $B$ and $\Theta$. To do this, we choose a threshold at which each element value is considered “zero” or “nonzero” and then we calculate the Precision (P) and Recall (R) for different synthetic data.
	
	(\romannumeral2) \textbf{Regression error of $B$ and $\Theta$}: We examine the prediction error of each model when using $\hat{\boldsymbol{B}}$ to predict $Y$ from $X$.
	
	For this analysis, we use a block-structured network over the outputs. The outputs are divided into non-overlapping groups, and each group forms a fully connected subgraph in the network.
	
	Here we describe our procedure for generating synthetic data. At a high level, we first fix the sparse structure of the underlying components of the model, then generate coefficient values, and lastly sample $X$ and $Y$ according to our model.
	
	Given the number of samples $n$, the number of genes $q$, and the number of SNPs $p$, firstly we determine the module size in the gene network and fix the number of SNPs $s$ associated with each gene. Next, we randomly assign each gene to a module and select the set of $s$ SNPs associated with each module.
	For each module, we randomly assign a major gene in this module and for each SNP $x_{j}$ associated with the primary gene $y_{k}$, we generate its association strengths according to $\beta_{jk} \sim Uniform(0,1)$. 
	For the other genes $y_{m}$ in this module, we generate the association strengths according to $\beta_{jm} \sim Uniform(\beta_{jk},\rho^{2})$, where $\rho = 0.1$ (we can change this parameter for different synthetic data). 
	
	Then, we consider four settings of the covariance matrices $E$ and $T$ to generate the simulated datasets. These are \textbf{case 1}: $T = I_{p×p}$ and $E = I_{q×q}$. \textbf{case 2}: $T = 0.6^{|j-k|}$ and $E = I_{q×q}$. \textbf{case 3}: $T = I_{p×p}$ and $E = 0.6^{|j-k|}$. \textbf{case 4}: $T = 0.6^{|j-k|}$ and $E = 0.6^{|j-k|}$.
	Finally, based on (\ref{e6}), we generate $\Theta = (E + B^{T}TB)^{-1}$.
	For each case, we average our results over 15 synthetic datasets.
	
\begin{figure}[H]
\centering
\subfigure[True]{
\includegraphics[width=4cm]{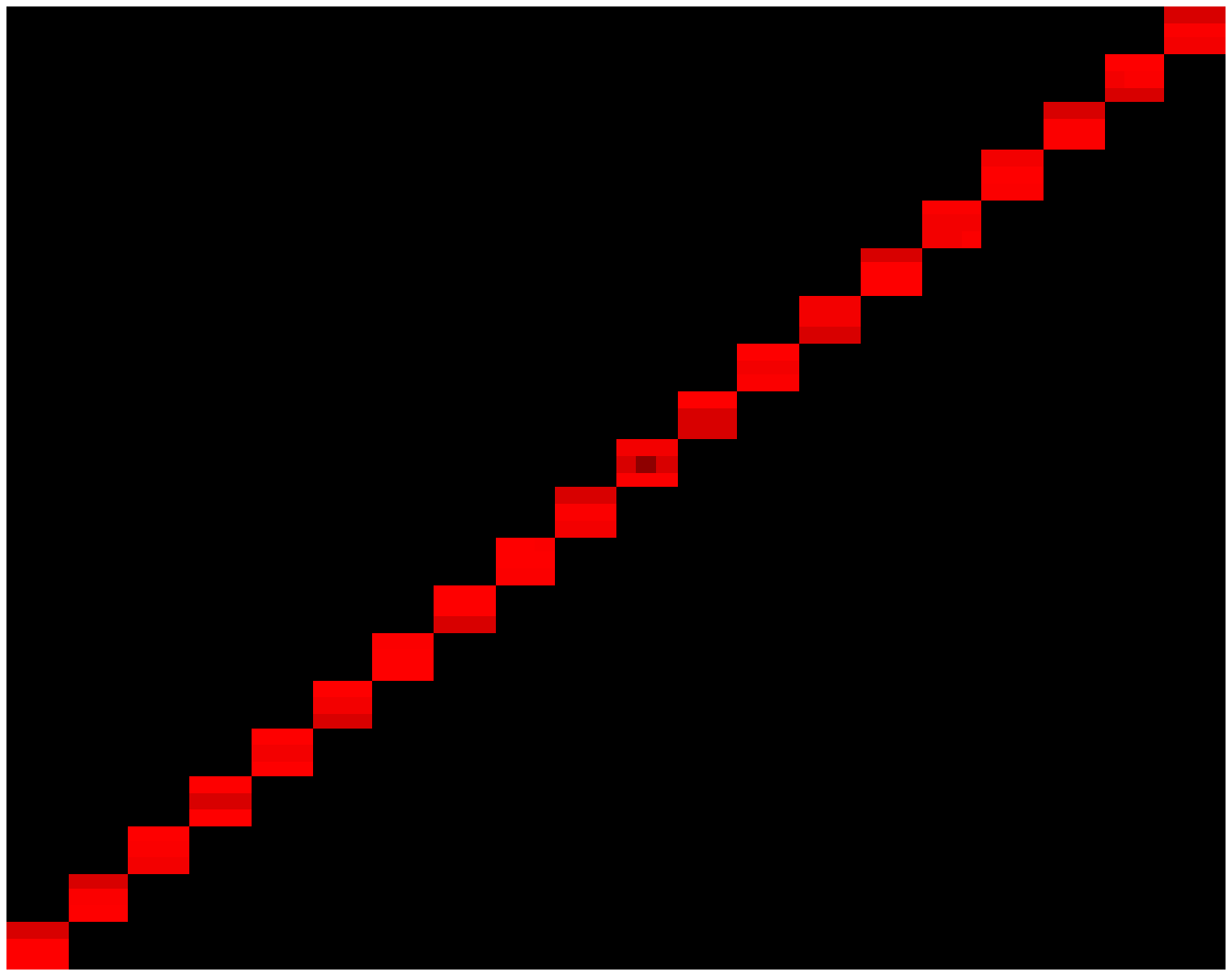}
}
\quad
\subfigure[GFLasso]{
\includegraphics[width=4cm]{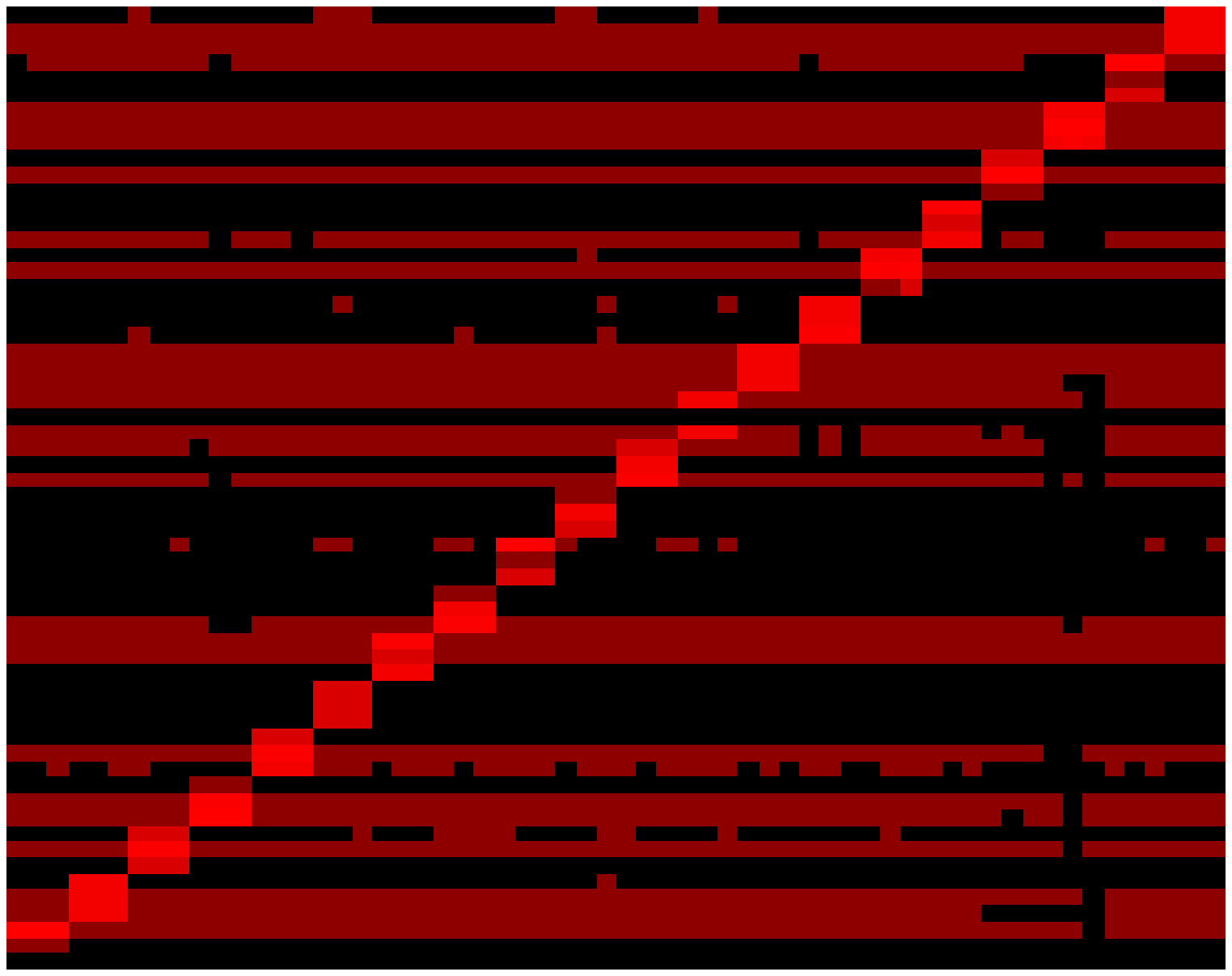}
}
\subfigure[MRCE]{
\includegraphics[width=4cm]{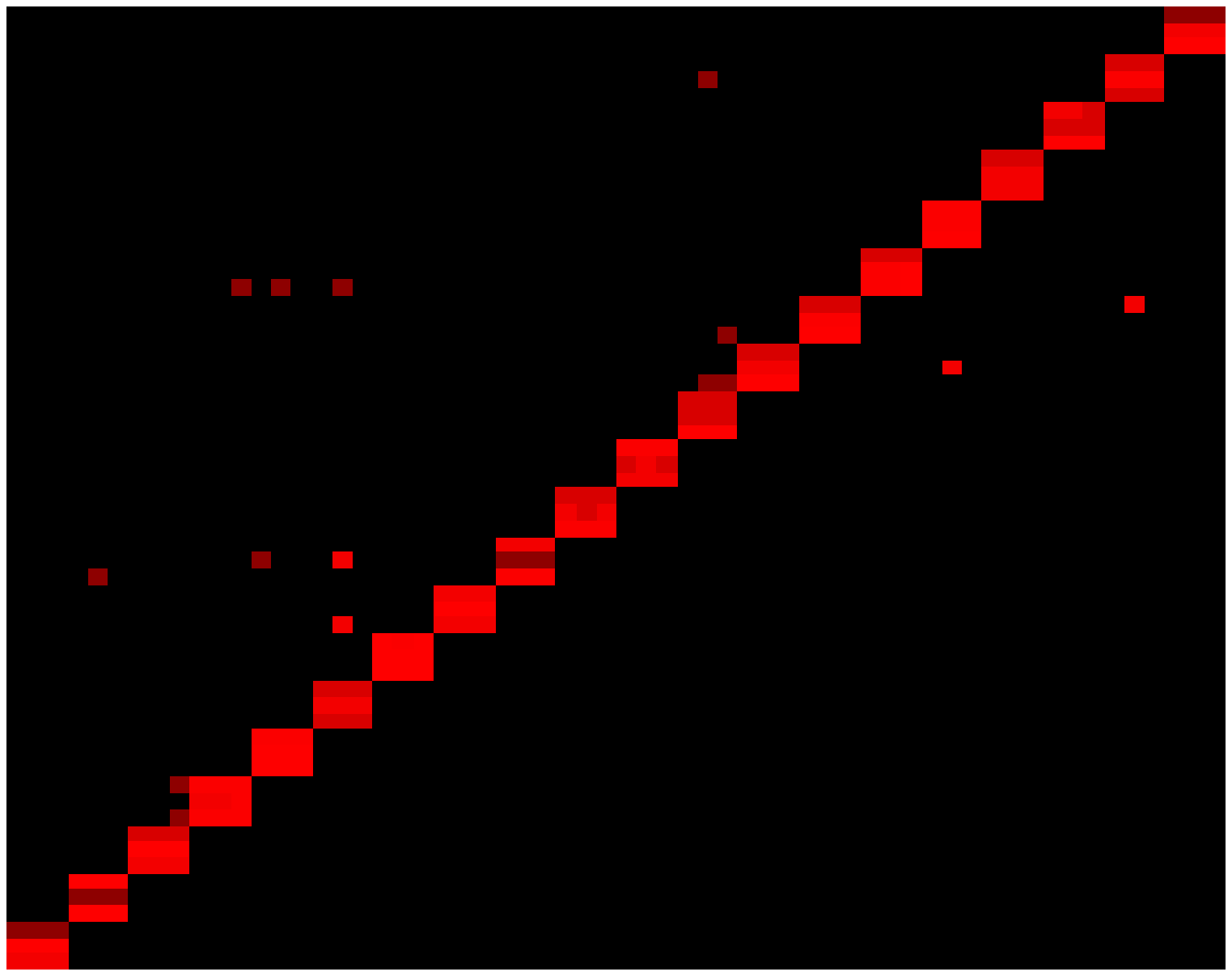}
}
\quad
\subfigure[ICLasso]{
\includegraphics[width=4cm]{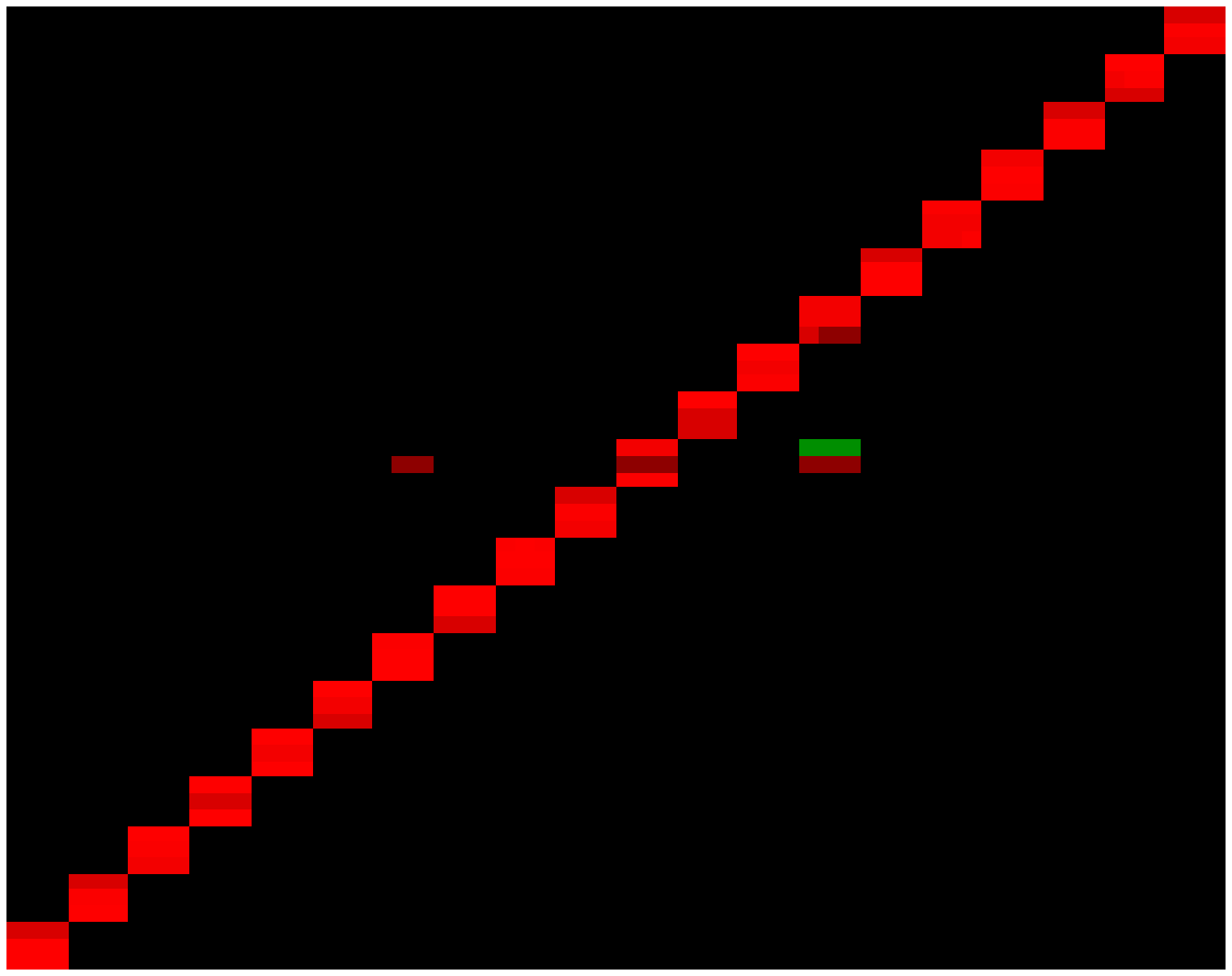}
}
\subfigure[$l_{1 \mbox{-} 2}$-GLasso]{
\includegraphics[width=4cm]{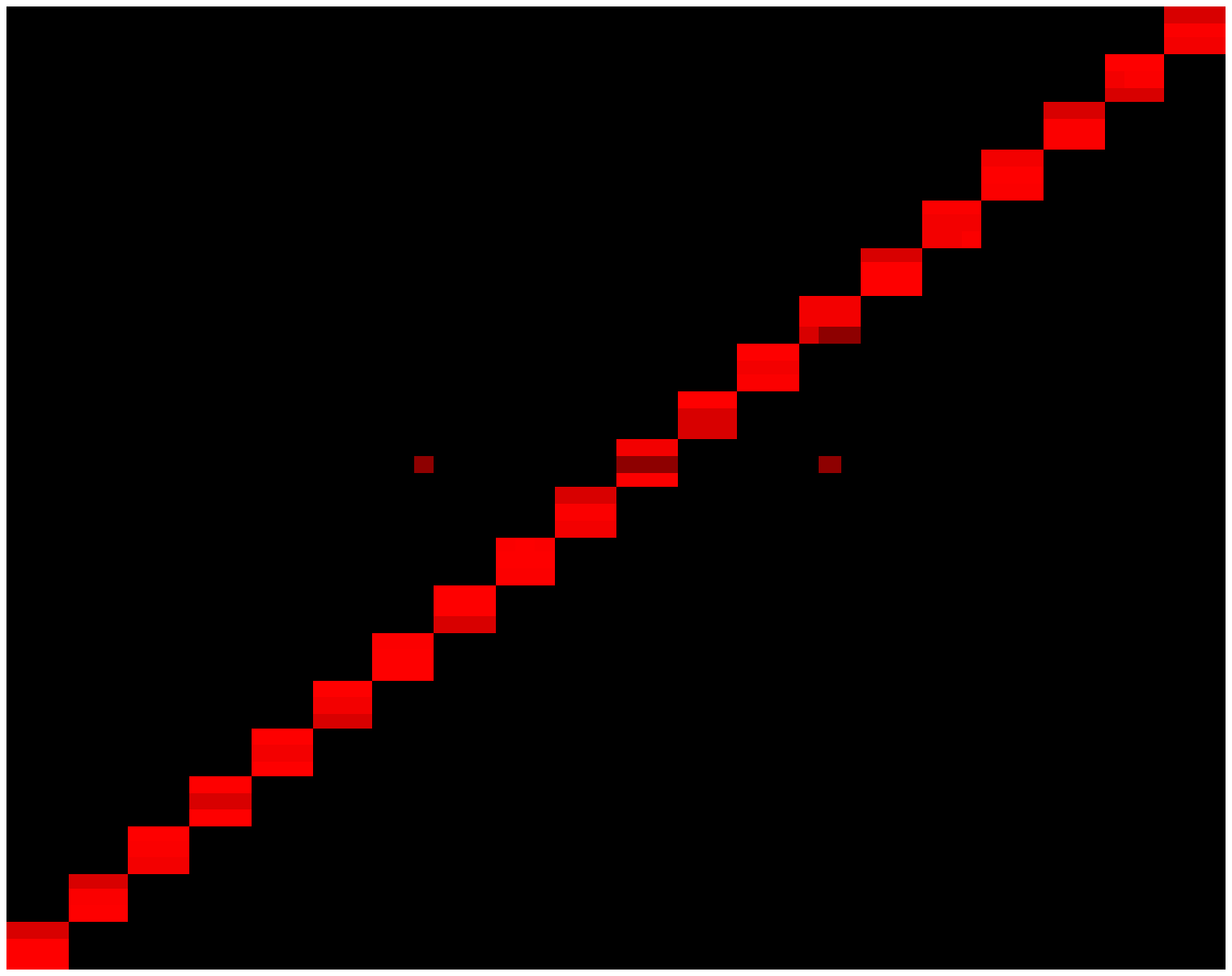}
}
\caption{The comparison of $B$ with different models on single synthetic dataset ($p = 60$ and $q = 60$). The real $B$ is given in the upper left corner of the figure.}
\label{f1}
\end{figure}

\begin{figure}[H]
\centering
\subfigure[True]{
\includegraphics[width=4cm]{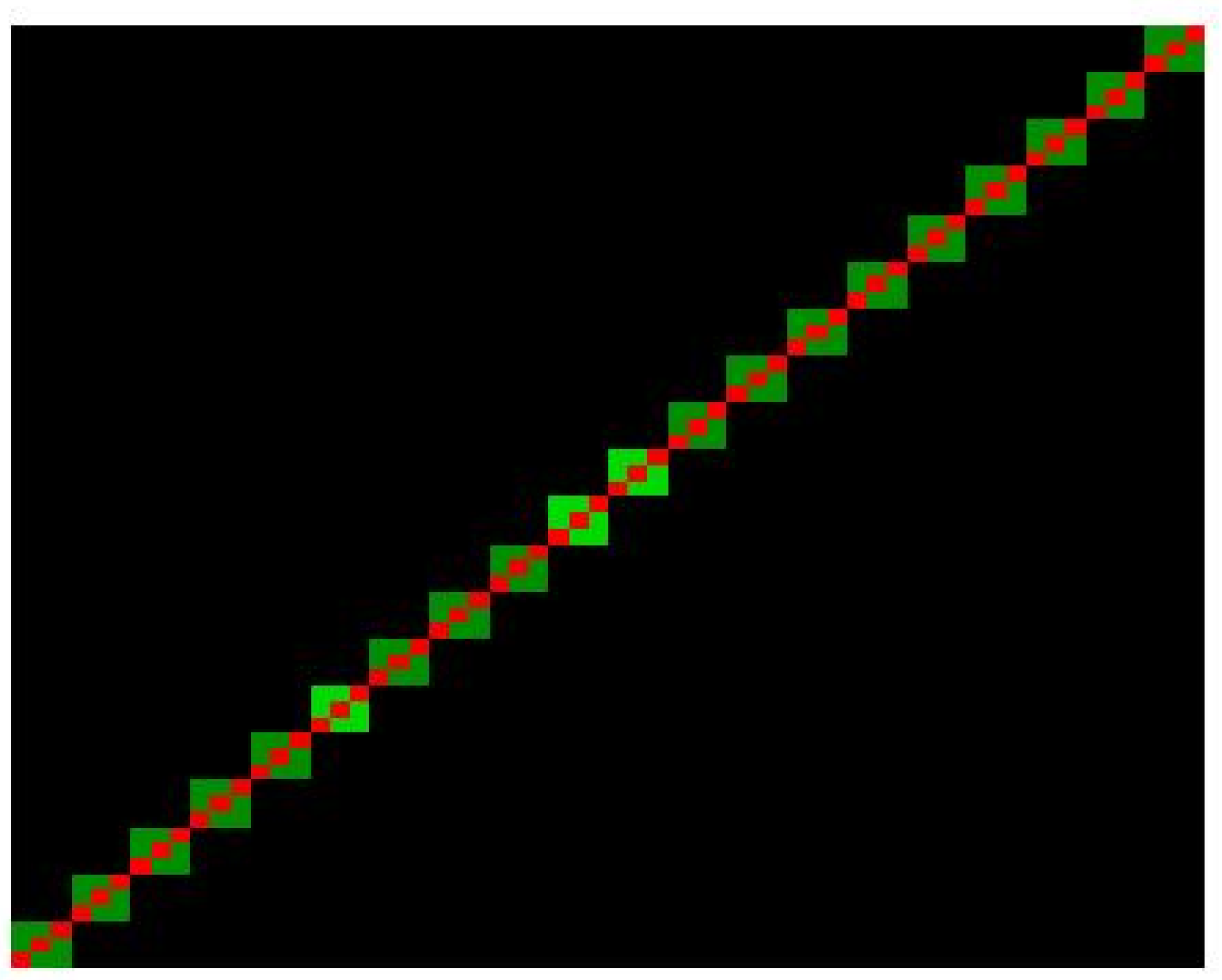}
}
\quad
\subfigure[GFLasso]{
\includegraphics[width=4cm]{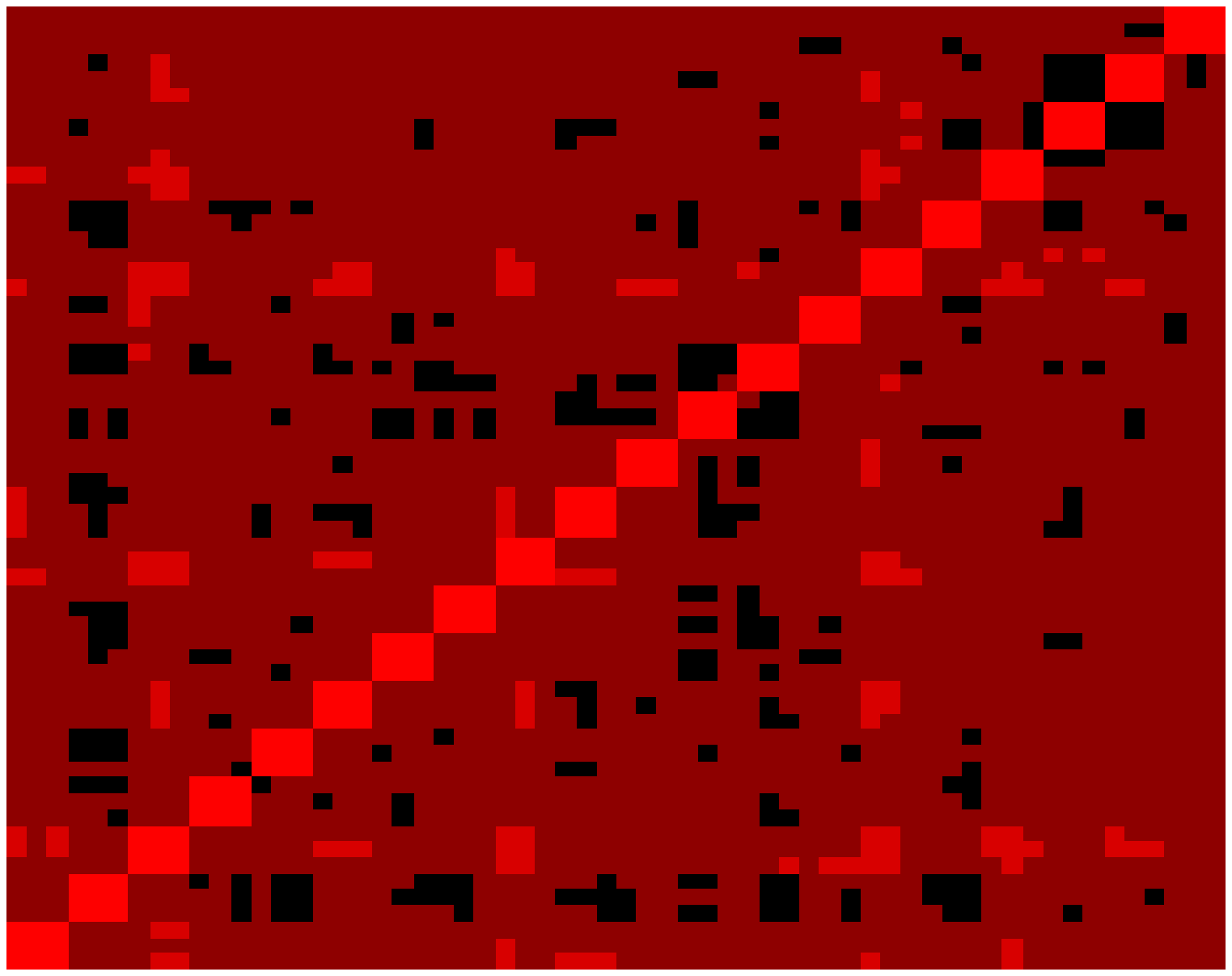}
}
\subfigure[MRCE]{
\includegraphics[width=4cm]{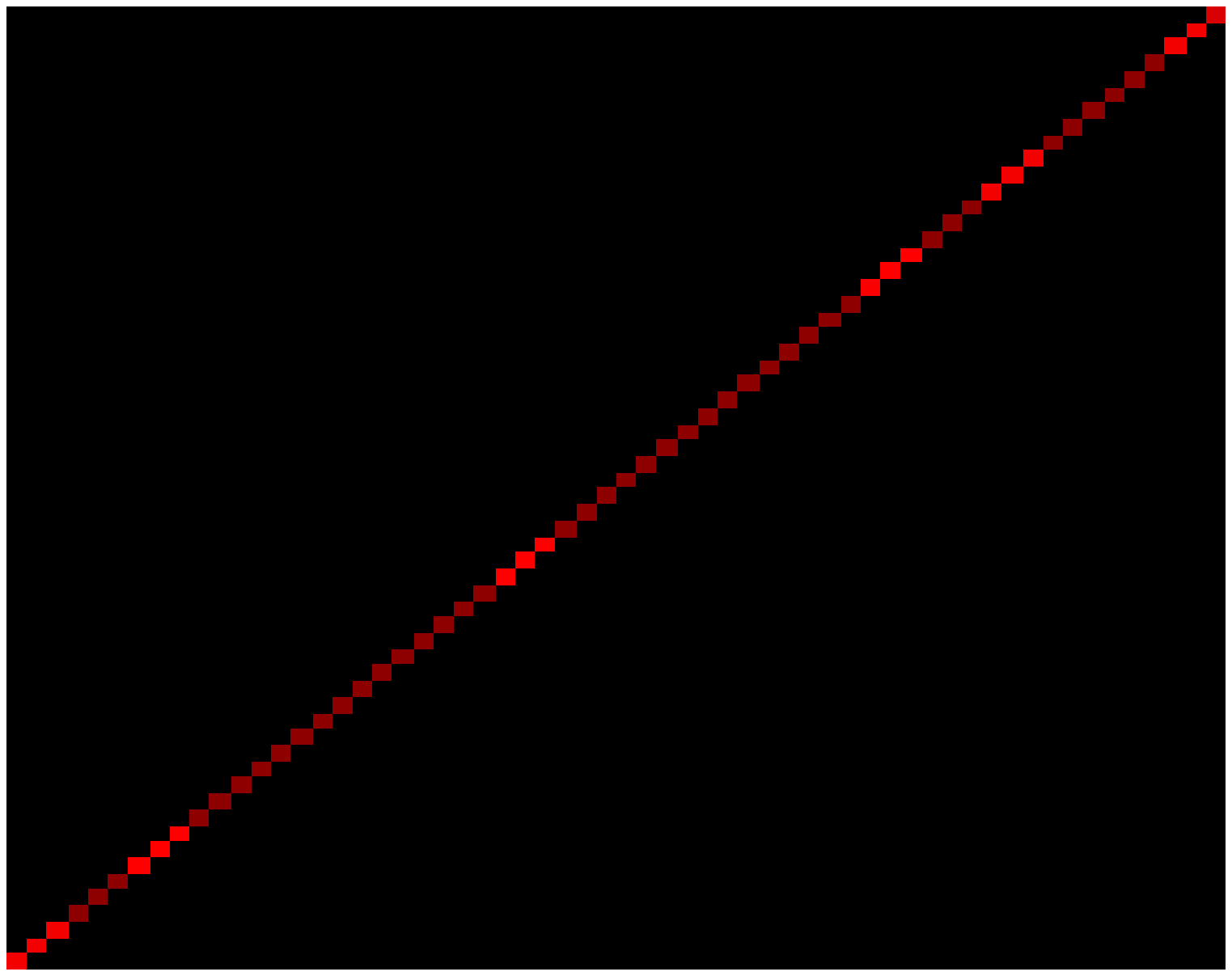}
}
\quad
\subfigure[ICLasso]{
\includegraphics[width=4cm]{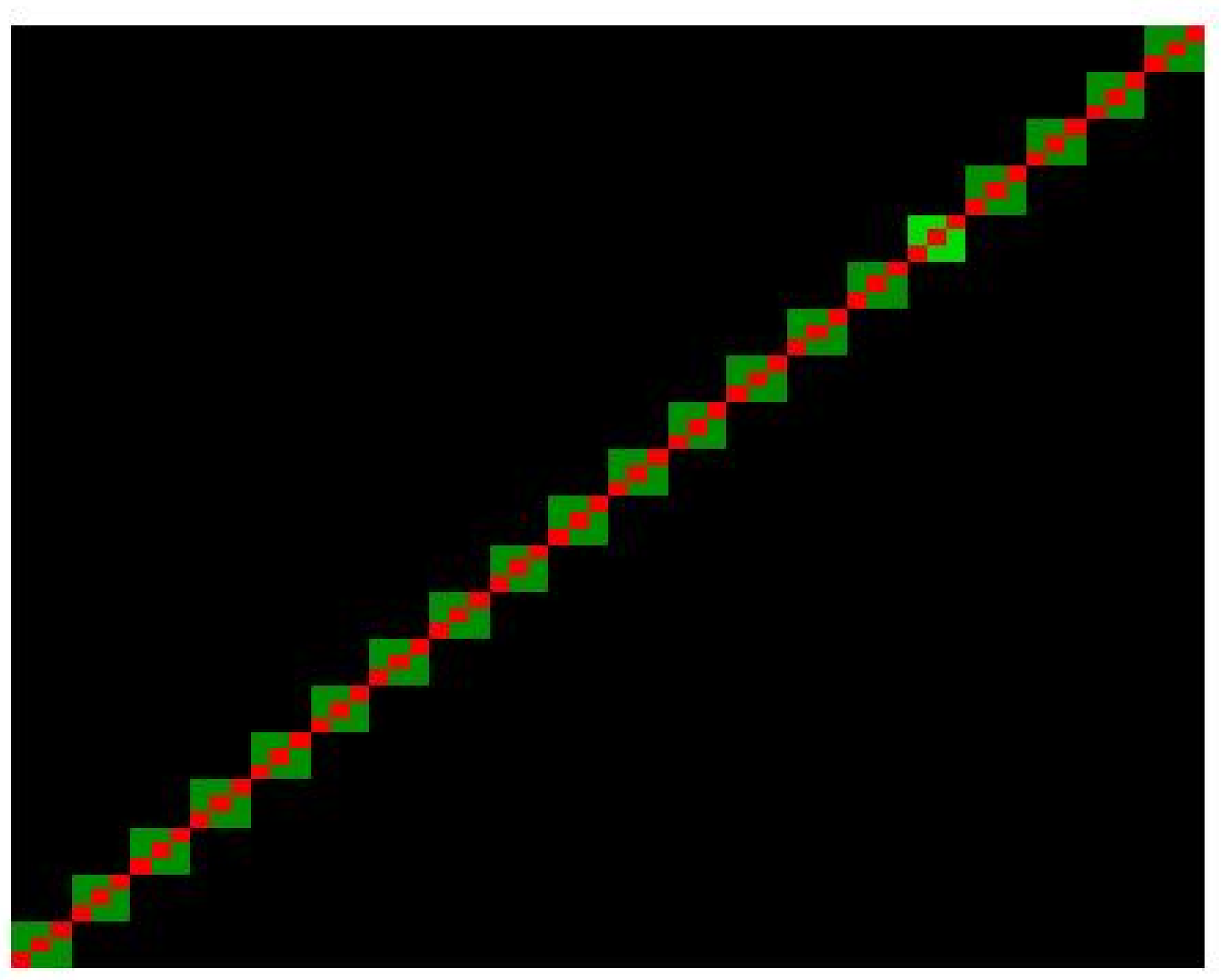}
}
\subfigure[$l_{1 \mbox{-} 2}$-GLasso]{
\includegraphics[width=4cm]{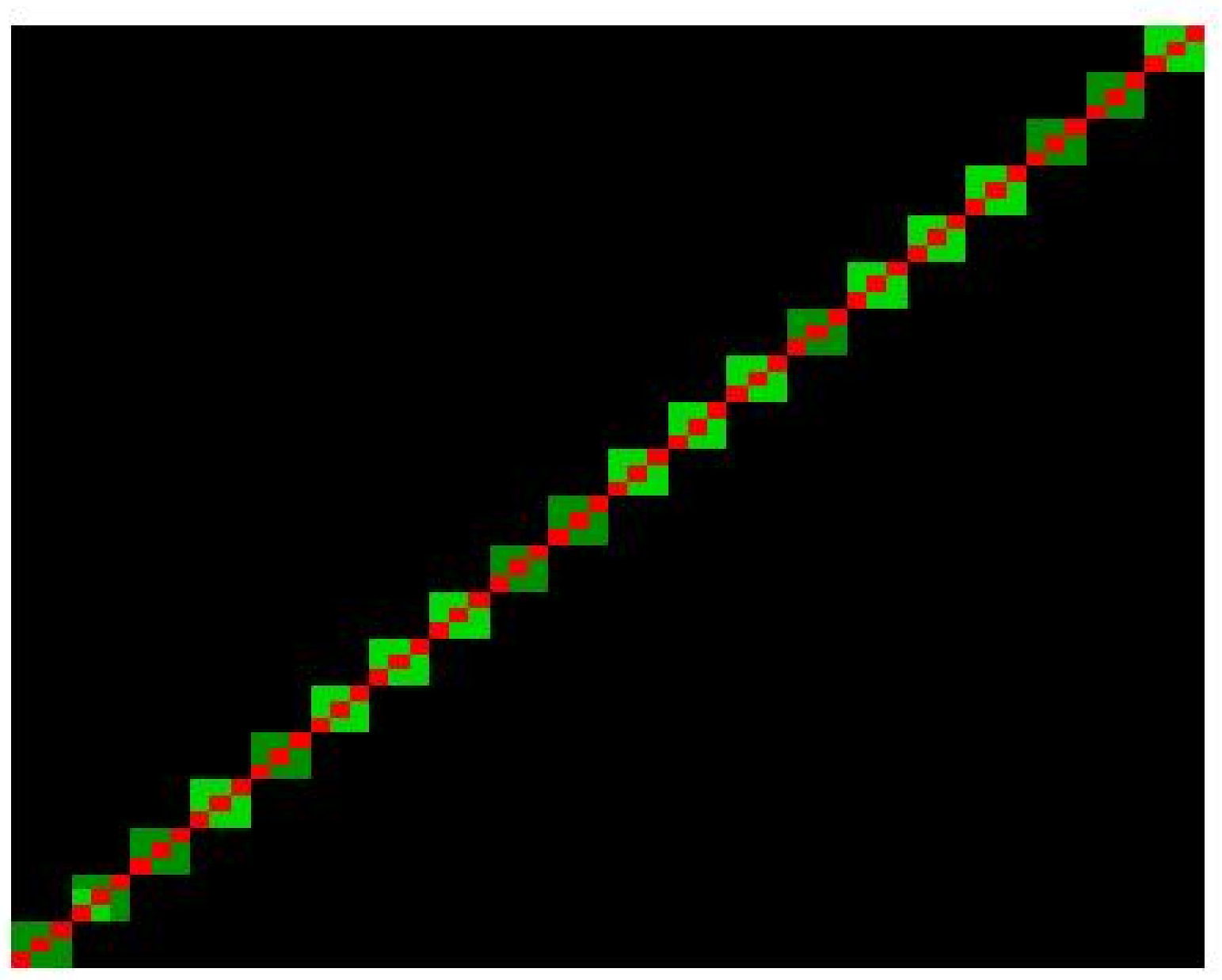}
}
\caption{The comparison of $\Theta$ with different models on single synthetic dataset ($p = 60$ and $q = 60$). The real $\Theta$ is given in the upper left corner of the figure.}
\label{f2}
\end{figure}

	The comparison of results on a single synthetic dataset with $n=120$, $p = 60$, $q = 60$ can be seen in Figure \ref{f1}, \ref{f2}. The detailed description is as follows: we identify 20 groups, and each group has 3 genes. In each group, $y_{1}$ is related to $x_{1}, x_{2},x_{3}$; $y_{4}$ is related to $x_{4},x_{5},x_{6}$ and so on. Let $E = I_{q×q}$ and $T \neq I_{p×p}$.
	
	The real $B$ and $\Theta$ are both given in the upper left corner of Figure \ref{f1}, \ref{f2}, the right side of Figure \ref{f1}, \ref{f2} show the results of GFLasso and MRCE and the bottom row shows the results of ICLasso and $l_{1 \mbox{-} 2}$-GLasso. All models can find the structure of $B$, but GFLasso's estimates are subject to significant error. By comparing with other models, it can be found that the regression coefficient matrix calculated by $l_{1 \mbox{-} 2}$-GLasso is closer to the real $B$. 

\begin{figure}[H]
\centering
\subfigure[case 1: $T = I_{p×p}$ and $E = I_{q×q}$]{
\includegraphics[width=6cm]{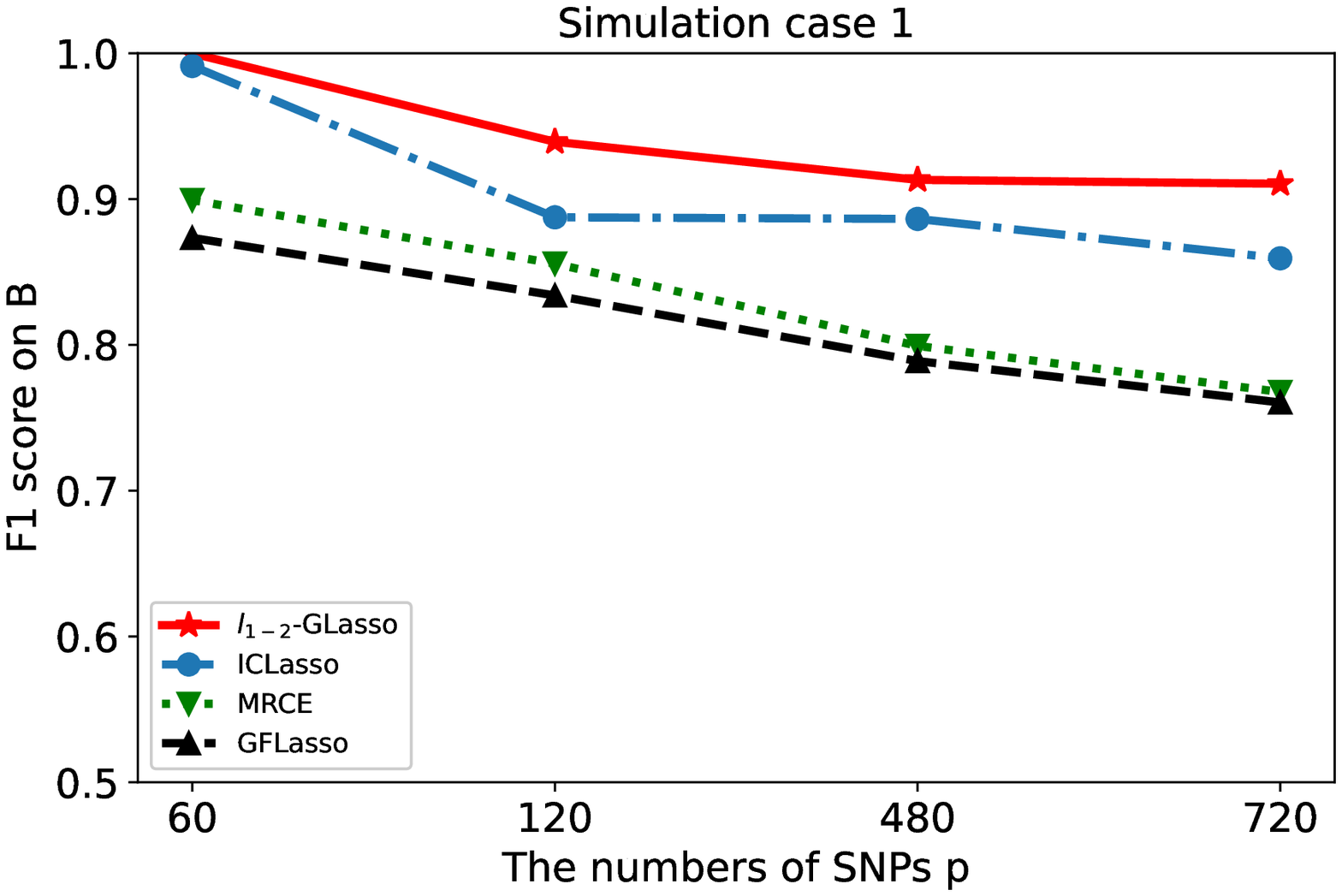}
}
\subfigure[case 2: $T = 0.6^{|j-k|}$ and $E = I_{q×q}$]{
\includegraphics[width=6cm]{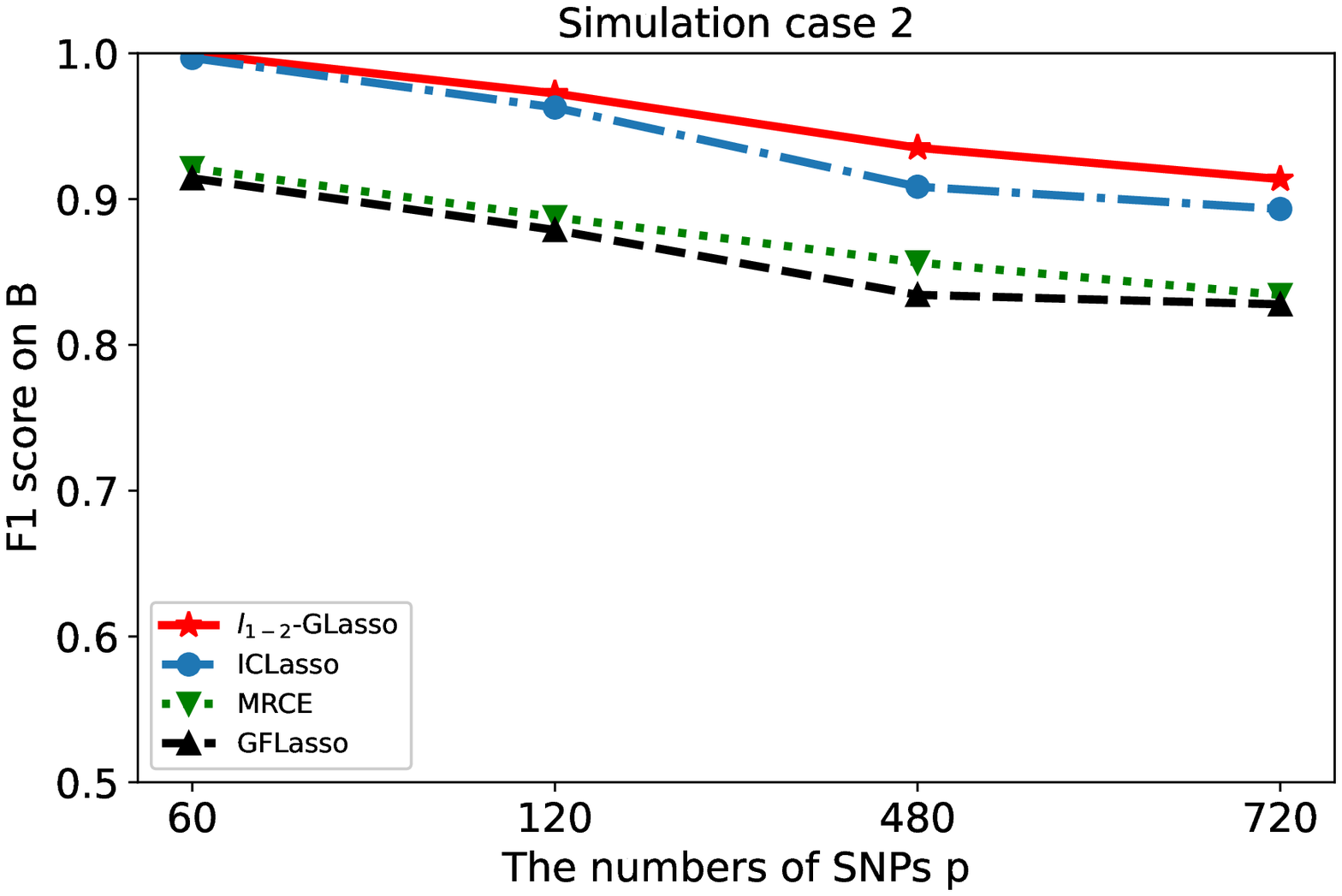}
}
\quad
\subfigure[case 3: $T = I_{p×p}$ and $E = 0.6^{|j-k|}$]{
\includegraphics[width=6cm]{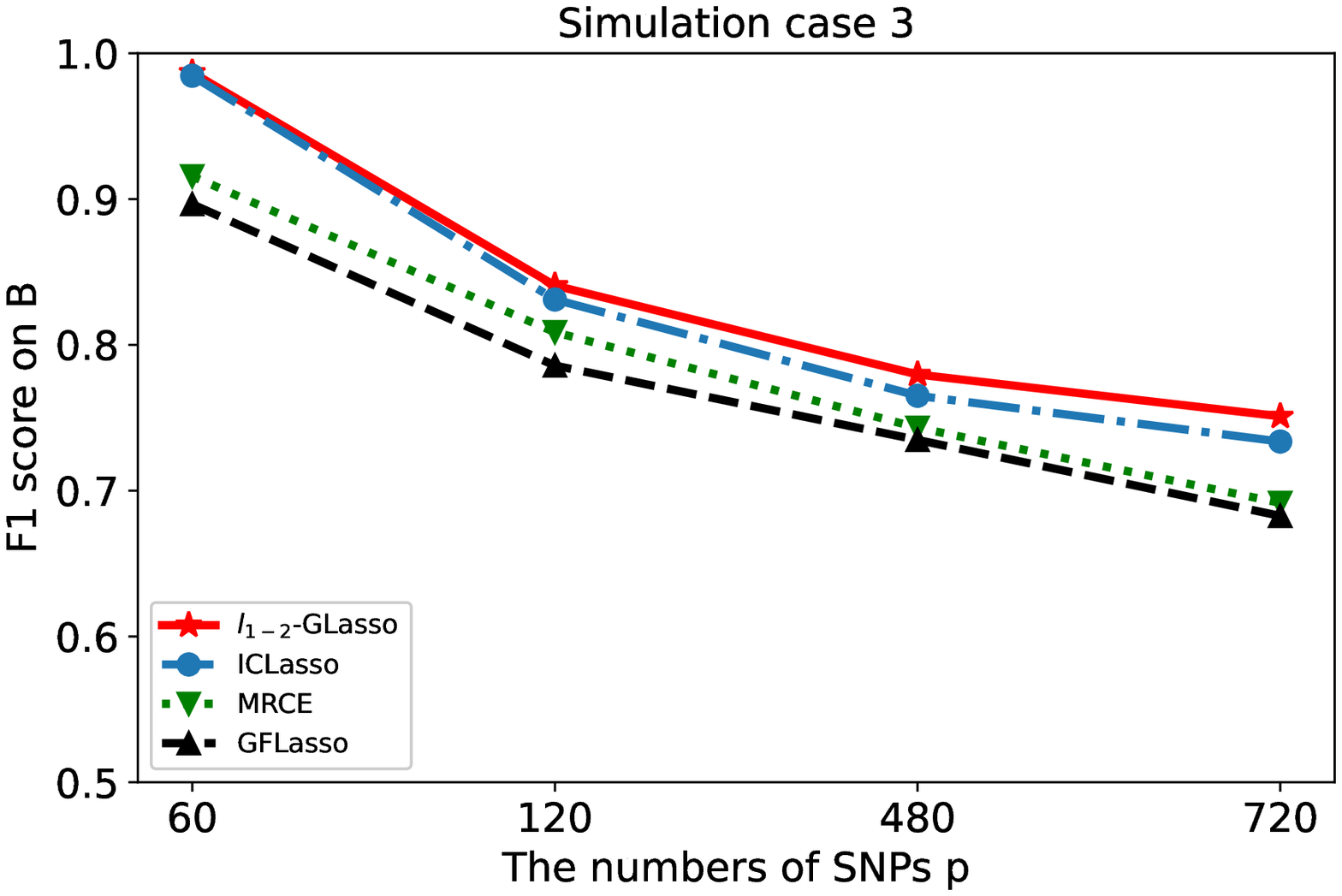}
}
\subfigure[case 4: $T = 0.6^{|j-k|}$ and $E = 0.6^{|j-k|}$]{
\includegraphics[width=6cm]{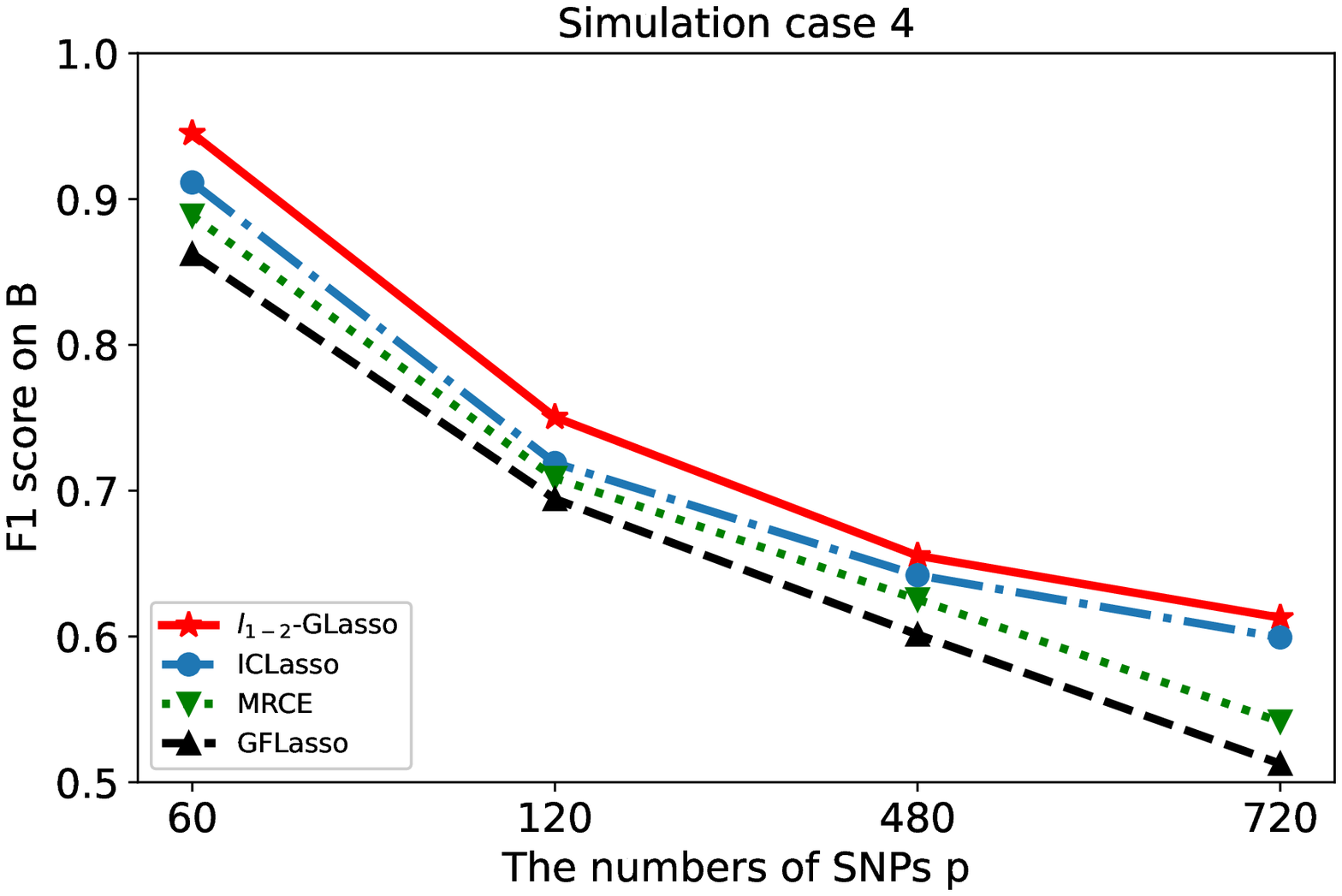}
}
\caption{The comparison of $B$ on 15 synthetic datasets generated with different types of covariance structures ($T$, $E$) and with different numbers of SNPs $p$.}
\label{f3}
\end{figure}

\begin{figure}[H]
\centering
\subfigure[case 1: $T = I_{p×p}$ and $E = I_{q×q}$]{
\includegraphics[width=6cm]{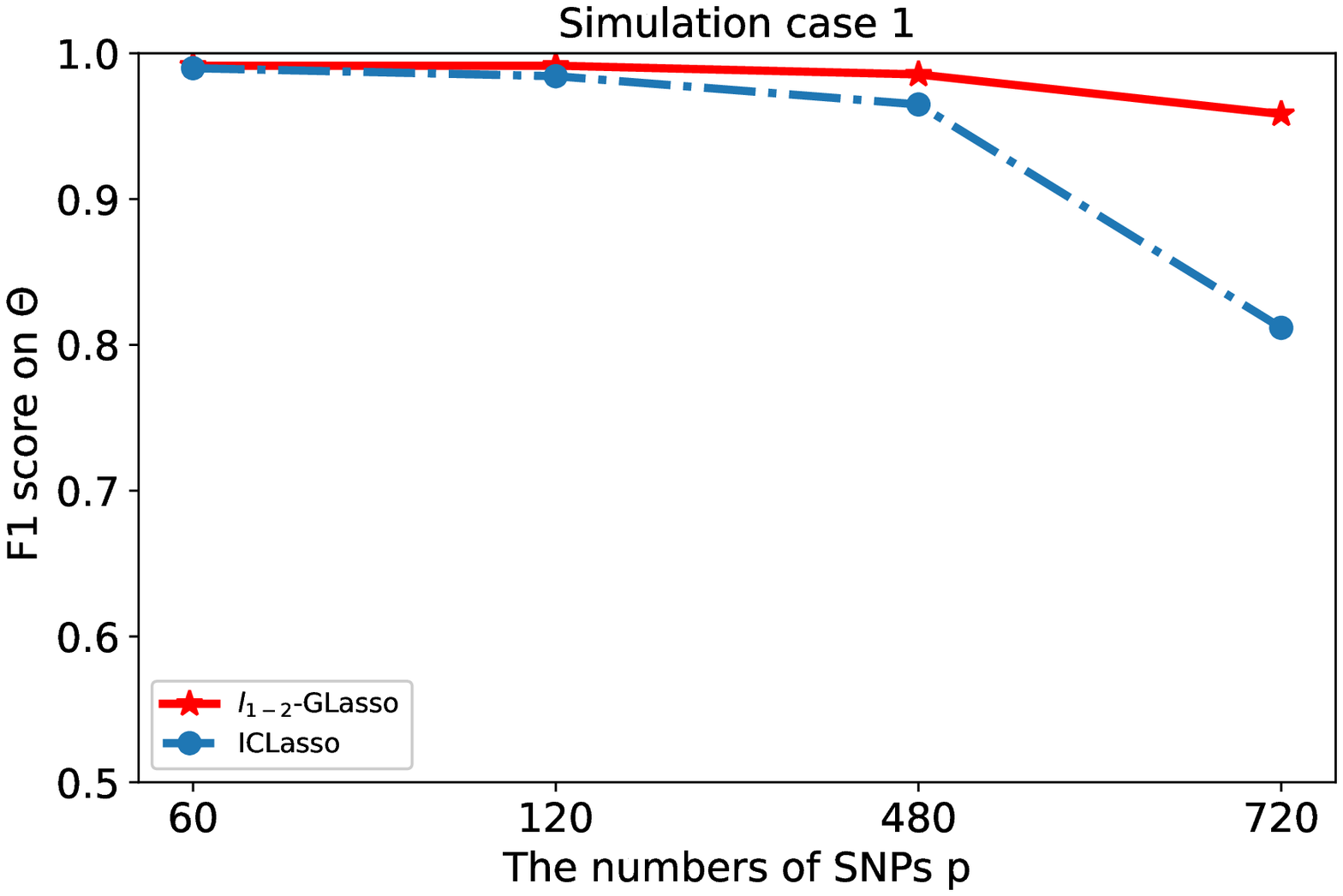}
}
\subfigure[case 2: $T = 0.6^{|j-k|}$ and $E = I_{q×q}$]{
\includegraphics[width=6cm]{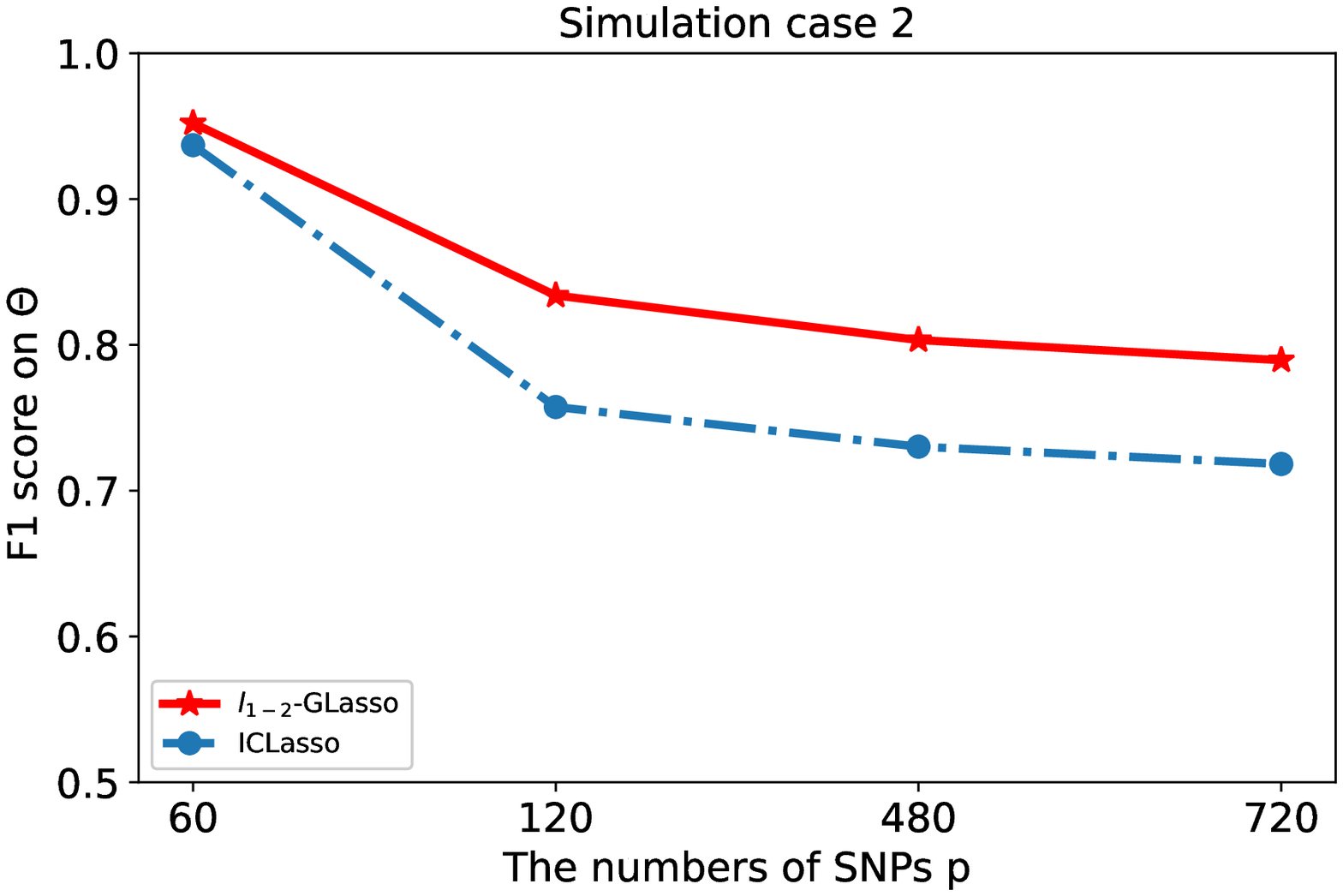}
}
\quad
\subfigure[case 3: $T = I_{p×p}$ and $E = 0.6^{|j-k|}$]{
\includegraphics[width=6cm]{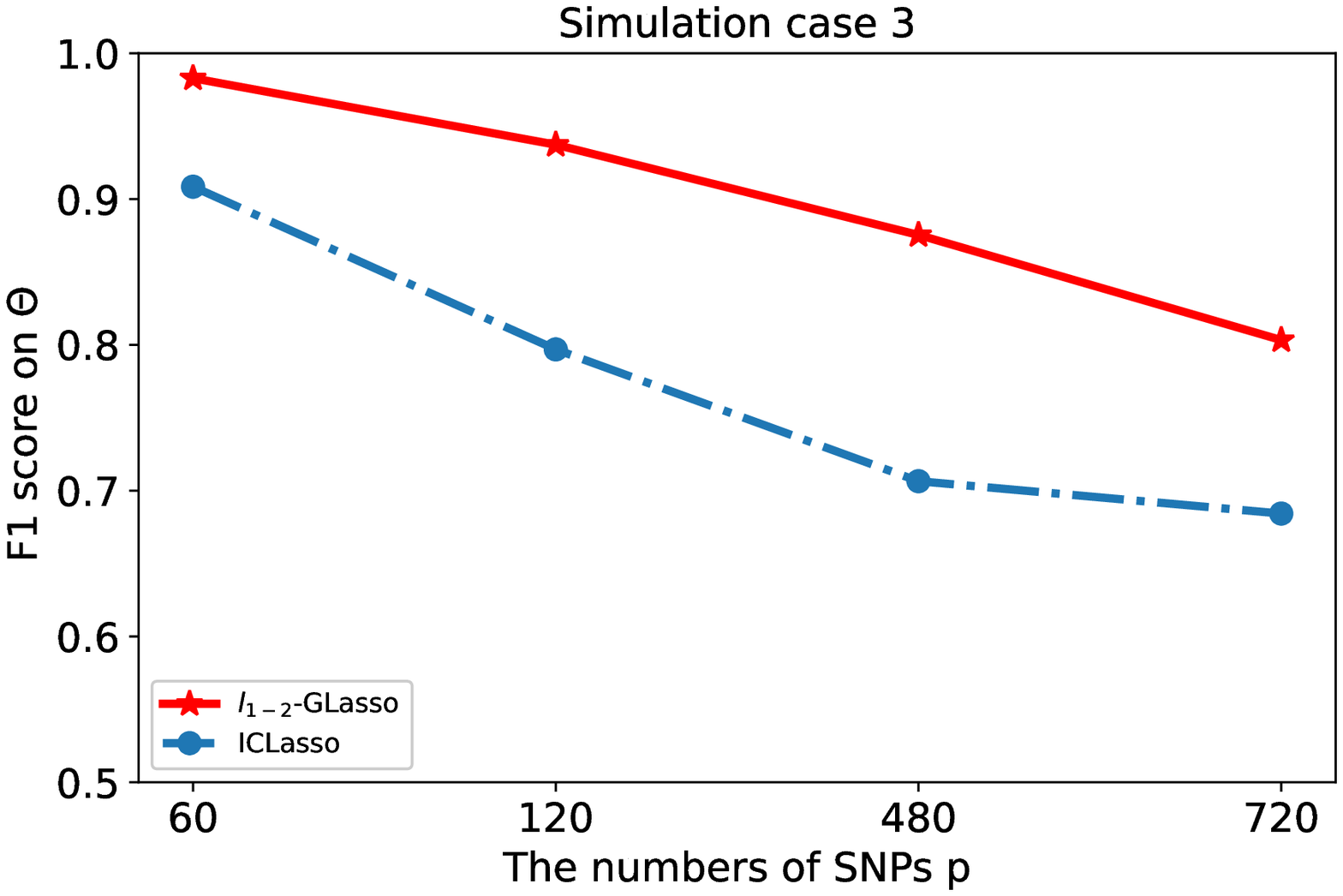}
}
\subfigure[case 4: $T = 0.6^{|j-k|}$ and $E = 0.6^{|j-k|}$]{
\includegraphics[width=6cm]{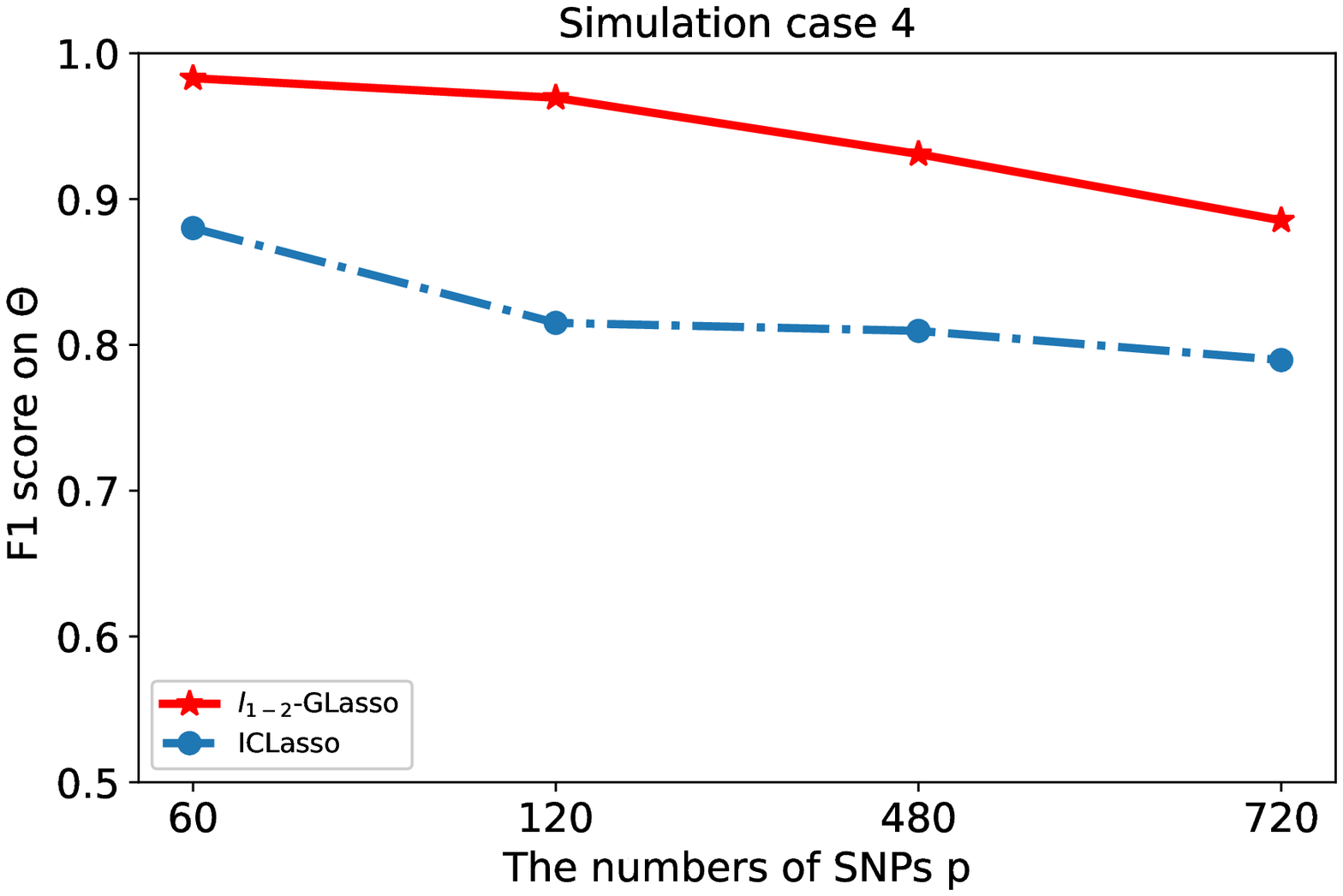}
}
\caption{The comparison of $\Theta$ on 15 synthetic datasets generated with different types of covariance structures ($T$, $E$) and with different numbers of SNPs $p$.}
\label{f4}
\end{figure}

    As can be seen from the Figure \ref{f2}, only $l_{1 \mbox{-} 2}$-GLasso and ICLasso can recover $\Theta$ more accurately.
	The main results of our synthetic experiments are shown in Figure \ref{f3}, \ref{f4}, \ref{f5}. We evaluate our approach according to three metrics: (1) F1 score on $B$ (2) F1 score on $\Theta$ (3) Regression error on $Y$. It can be seen that $l_{1 \mbox{-} 2}$-GLasso is closer to the real data in terms of sparsity. Comparing the size of different input dimensions, $l_{1 \mbox{-} 2}$-GLasso is also superior to other models.


\begin{figure}[H]\label{f5}
\centering
\subfigure[case 1: $T = I_{p×p}$ and $E = I_{q×q}$]{
\includegraphics[width=6cm]{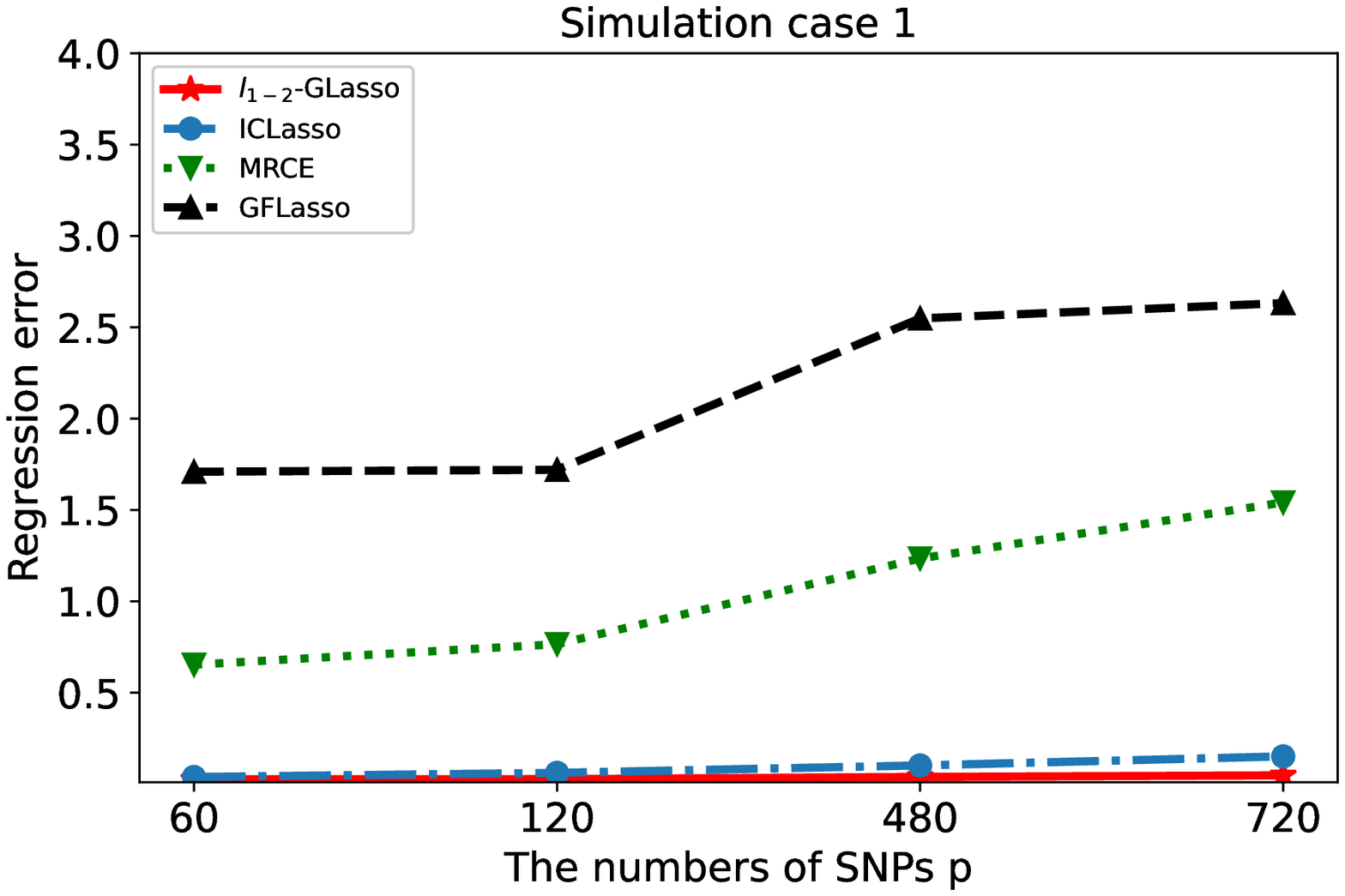}
}
\subfigure[case 2: $T = 0.6^{|j-k|}$ and $E = I_{q×q}$]{
\includegraphics[width=6cm]{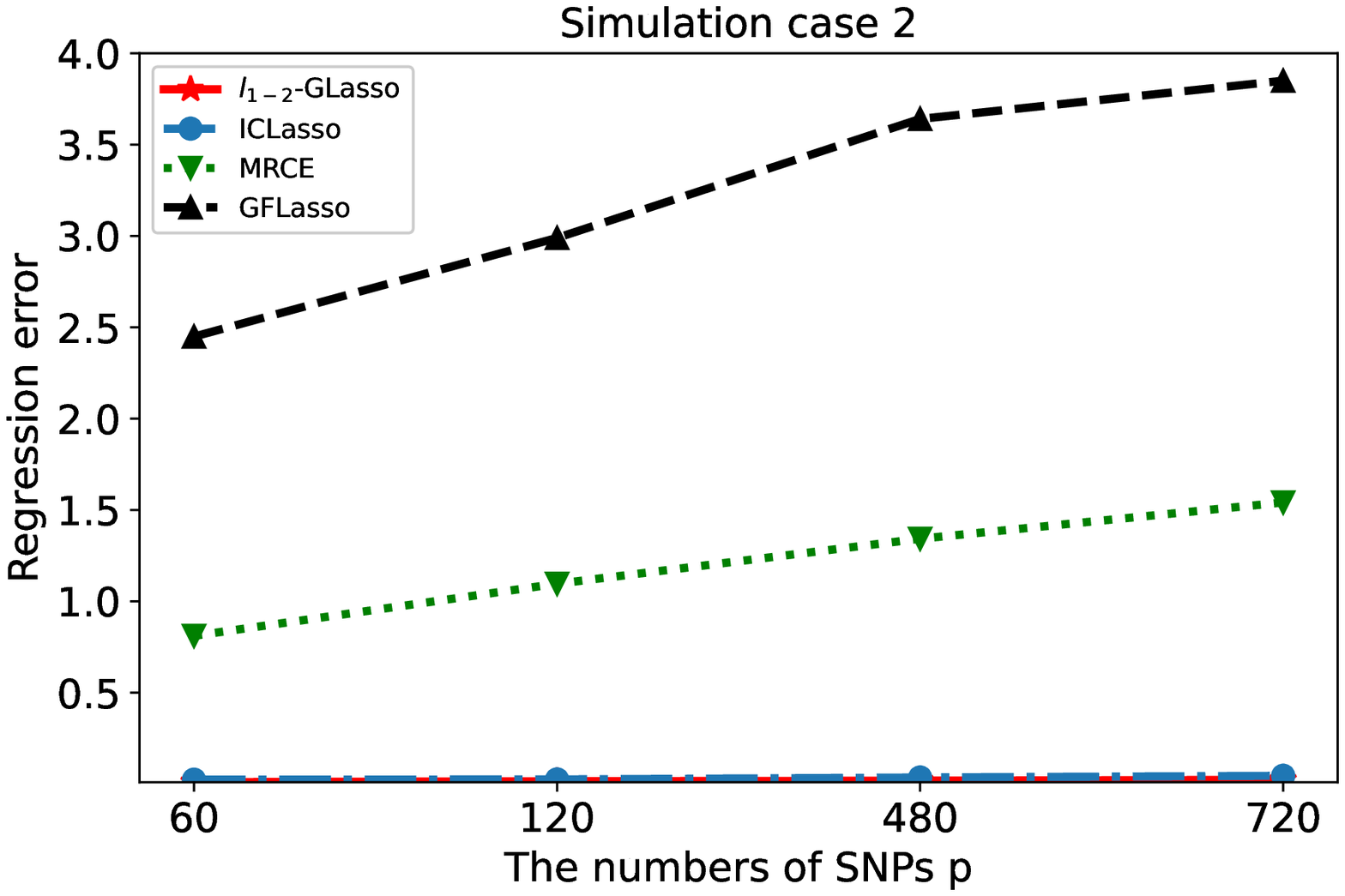}
}
\quad
\subfigure[case 3: $T = I_{p×p}$ and $E = 0.6^{|j-k|}$]{
\includegraphics[width=6cm]{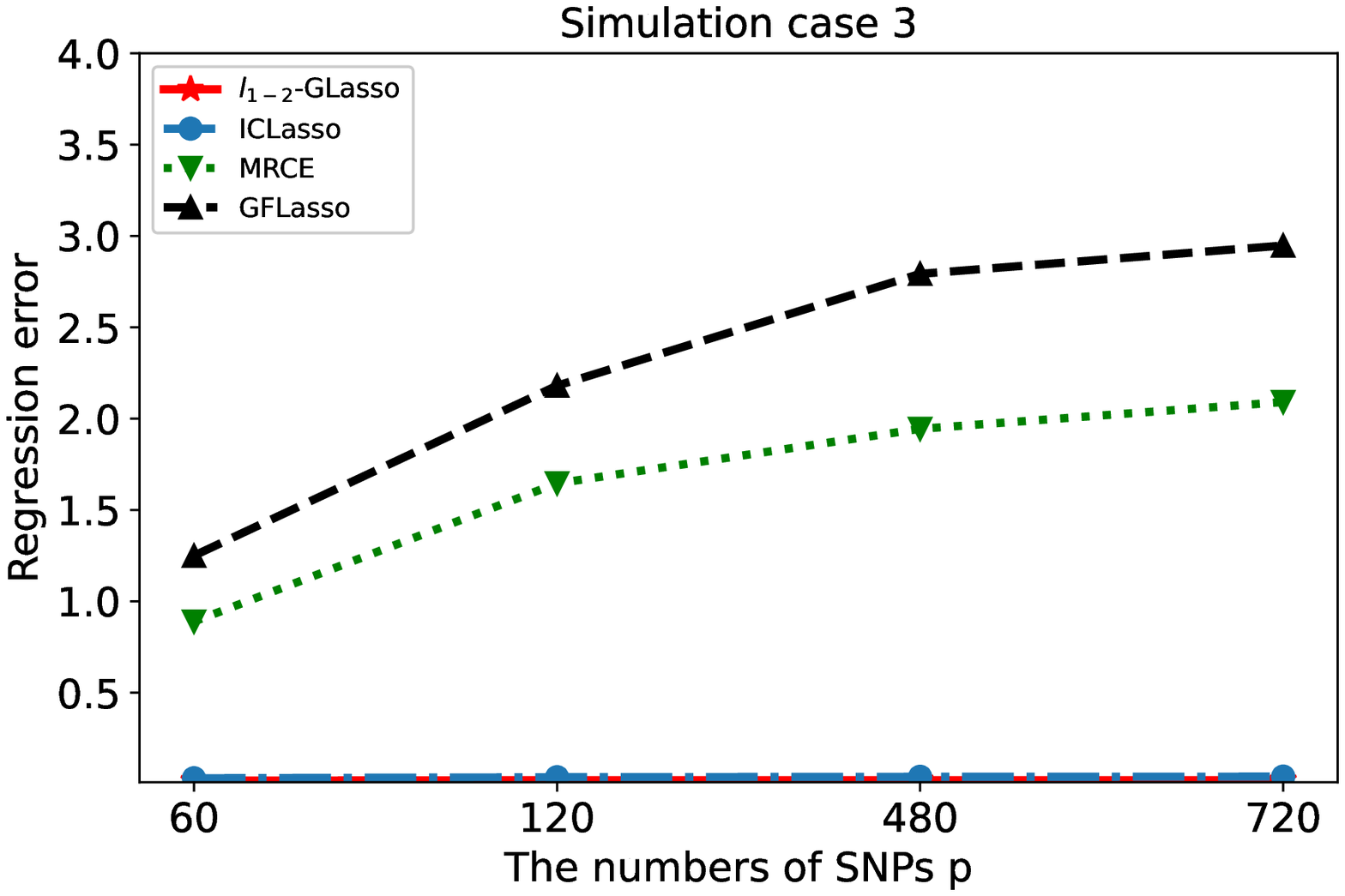}
}
\subfigure[case 4: $T = 0.6^{|j-k|}$ and $E = 0.6^{|j-k|}$]{
\includegraphics[width=6cm]{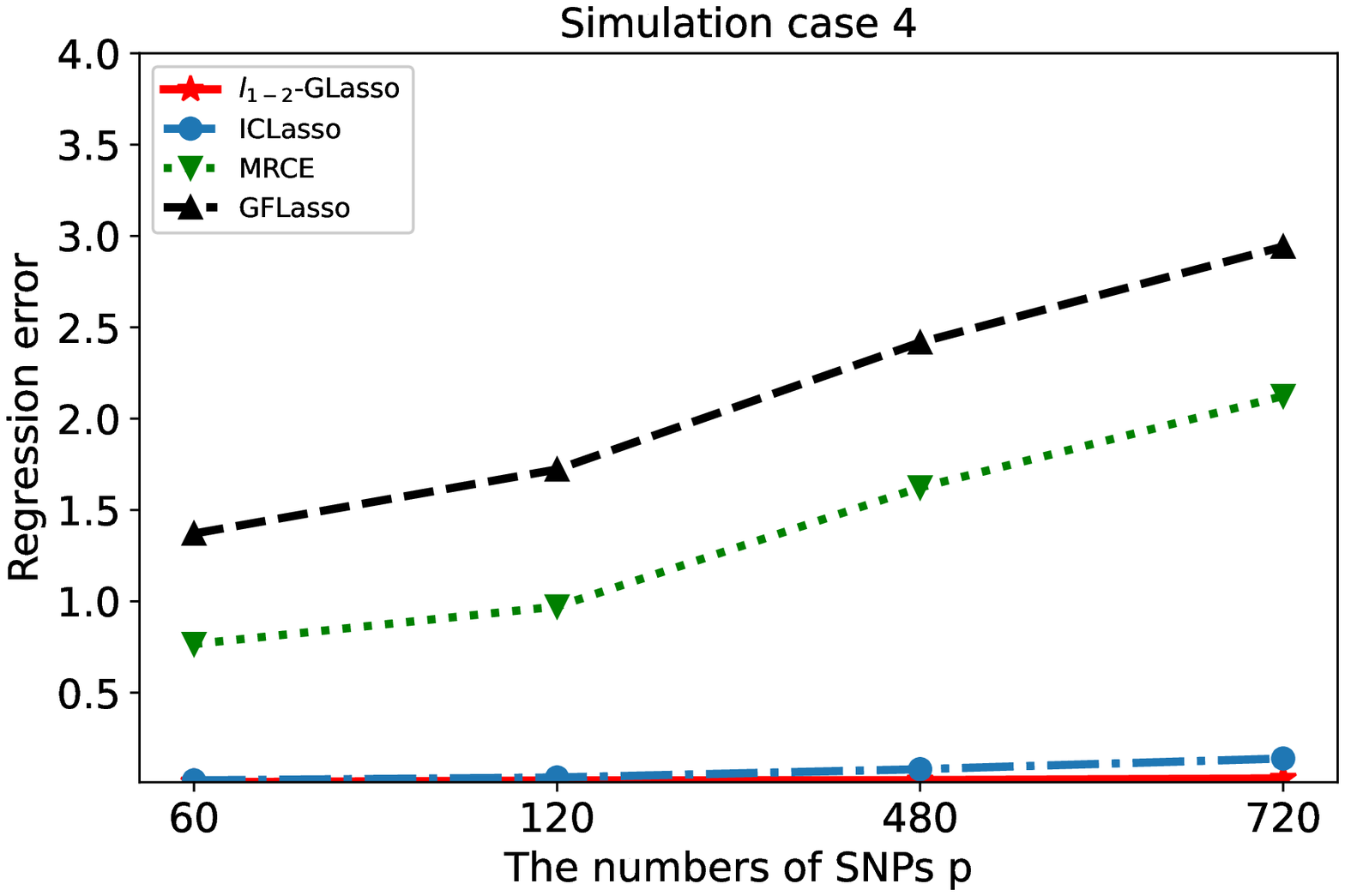}
}
\caption{The comparison of Regression Error on 15 synthetic datasets generated with different types of covariance structures ($T$, $E$) and with different numbers of SNPs $p$.}
\label{f5}
\end{figure}
In Figure \ref{f3}, \ref{f4}, we present the F1 score in the recovery of $B$ and $\Theta$. To some extent, the F1 score reflects the ability of each model to learn the regression coefficient matrix and the output structure. In Figure \ref{f5}, we show the regression error of $Y$. It can be seen that in terms of regression error, $l_{1 \mbox{-} 2}$-GLasso can be a little better than ICLasso. As shown in these Figures, our results clearly show that $l_{1 \mbox{-} 2}$-GLasso outperforms other baselines in the four cases we considered.

\begin{figure}[H]
\centering
\includegraphics[scale=0.6]{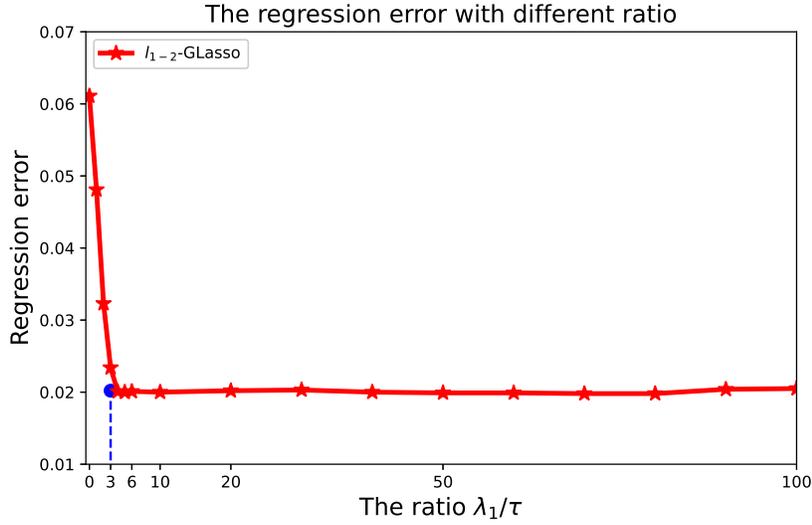}
\caption{The regression error with different ratio $\frac{\lambda_{1}}{\tau}$}
\label{f6}
\end{figure}

Finally, because $l_{1 \mbox{-} 2}$-GLasso introduces a new hyperparameter $\tau$, we also explore the influence of different hyperparameters on the effect of our model. We use the dataset shown in Figure \ref{f1}, and  the regression error with different ratios $\frac{\lambda_{1}}{\tau}$ is presented in Figure \ref{f6}. As we can see, when $\frac{\lambda_{1}}{\tau} < 3$, the regression error is minimized. A larger ratio leads to a more satisfying solution. When $\frac{\lambda_{1}}{\tau} \geq 3$, the regression error becomes more and more stable, so we finally select $\frac{\lambda_{1}}{\tau} = 10$ to conduct the following experiments.

\section{Real dataset From eQTL Mapping}
\cite{greenlaw2017bayesian} analyze a dataset obtained from the ADNI-1 database. We compare $l_{1 \mbox{-} 2}$-GLasso with other models on this dataset. The genes used in our analysis are listed in Table \ref{t1}. 

\begin{table}[H]
\begin{center} 
\caption{The Gene ID in the ADNI-1 database}
\begin{tabular}{llll}
\hline
\multicolumn{1}{l}{Gene ID} & \multicolumn{1}{l}{Measurement} & \multicolumn{1}{l}{Region of interest}\\ \hline
Left\_AmygVol & Volume & Amygdala\\
Left\_CerebCtx & Volume & Cerebral cortex\\
Left\_CerebWM & Volume &  Cerebral white matter\\
Left\_HippVol & Volume &  Hippocampus\\
Left\_InfLatVent & Volume &  Inferior lateral ventricle\\
Left\_LatVent & Volume & Lateral ventricle\\
Left\_EntCtx & Thickness & Entorhinal cortex\\
Left\_Fusiform & Thickness &  Fusiform gyrus\\
Left\_InfParietal & Thickness &  Inferior parietal gyrus\\
Left\_InfTemporal & Thickness & Inferior temporal gyrus \\
Left\_MidTemporal & Thickness &  Middle temporal gyrus\\
Left\_Parahipp & Thickness & Parahippocampal gyrus\\
Left\_PostCing & Thickness &  Posterior cingulate\\
Left\_Postcentral & Thickness &  Postcentral gyrus\\
Left\_Precentral & Thickness & Precentral gyurs\\ \hline
\label{t1}
\end{tabular}
\end{center}
\end{table}
The dataset is available for $n = 632$ subjects, and among all possible SNPs, we include only those SNPs belonging to the top 15 candidate genes listed on the AlzGene database. The dataset presented here is queried from the most recent genome build as of December 2014, from the ADNI-1 database. After quality control and imputation steps, the genetic dataset used for this study includes $p = 486$ SNPs from $q = 15$ genes. 
 
\begin{figure}[H]
\centering
\includegraphics[scale=0.8]{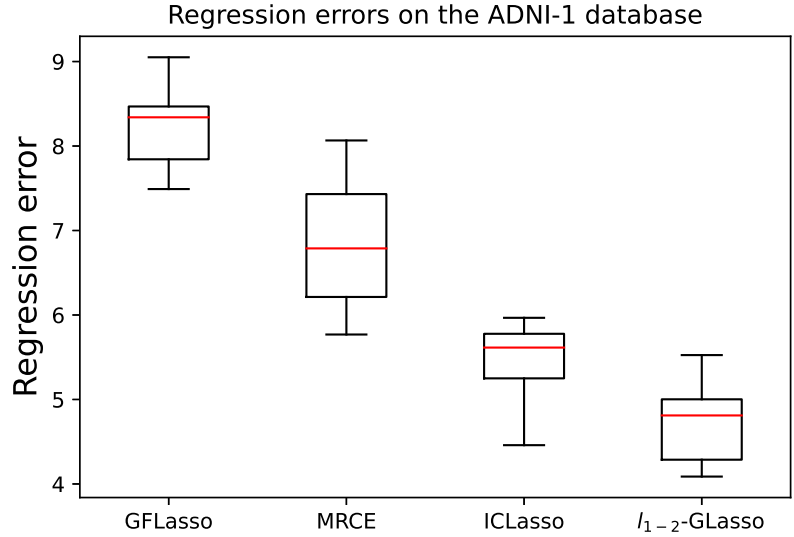}
\caption{Each box shows the discrete distribution of the regression error on these 10 different datasets. The median is shown by the red line.}
\label{f7}
\end{figure}

We apply $l_{1 \mbox{-} 2}$-GLasso and other models to discover SNPs that influence the expression levels of genes. This type of study is widely known as an expression quantitative trait locus (eQTL) mapping in the genetics community. It is generally believed that when a genetic variation in the genome such as an SNP perturbs the expression of a gene, the effect propagates through the gene network to influence the expressions of genes in downstream of the pathway. 
First, we estimate the regression coefficient matrix using 480 samples and then compute the regression error on the remaining 152 samples. As shown in Figure \ref{f7}, $l_{1 \mbox{-} 2}$-GLasso produces the smallest regression error. It should be noted that we randomly select 480 samples each time and repeat the sample 10 times. We show the median (red) and the discrete distribution of the regression error on these 10 different datasets.

In the original methodology of \cite{wang2012identifying}, based on biological experiments, 24 SNPs that are highly correlated with 15 genes have been detected. 
According to the results of biological experiments, we also apply the proposed model to find related SNPs. $l_{1 \mbox{-} 2}$-GLasso finds 22 SNPs among the determined 24 SNPs and these 22 SNPs are highlighted in Table \ref{tab5}. In addition, we also compared the GFLasso, MRCE, and ICLasso on the dataset with our model in Table \ref{t2}, \ref{t3}, \ref{t4}. The SNPs in bold are selected by each model. It can be seen in Table \ref{t4}, \ref{tab5} that $l_{1 \mbox{-} 2}$-GLasso and ICLasso can both estimate the structure of two larger subnetworks in the gene networks. This has been demonstrated by biological experiments. They show that two SNPs: rs405509 and rs10787010 stand out as being potentially associated with the largest number of ROIs. In Table \ref{t6}, we show the SNPs associated with genes identified by different models. $l_{1 \mbox{-} 2}$-GLasso is better than the other models.
To show the superiority of $l_{1 \mbox{-} 2}$-GLasso, we can see from Table \ref{t2}, \ref{t3}, \ref{t4}, GFLasso picks out 14 of the determined 24 SNPs , MRCE picks out 16 of the determined 24 SNPs and ICLasso picks out 17 of the determined SNPs. The experimental results are in line with our assumptions about $l_{1 \mbox{-} 2}$-GLasso. Compared with other models, $l_{1 \mbox{-} 2}$-GLasso can more accurately determine the SNPs related to genes.  

\begin{figure}[H]
\centering
\subfigure[$B$ calculated by ICLasso]{
\includegraphics[width=6cm]{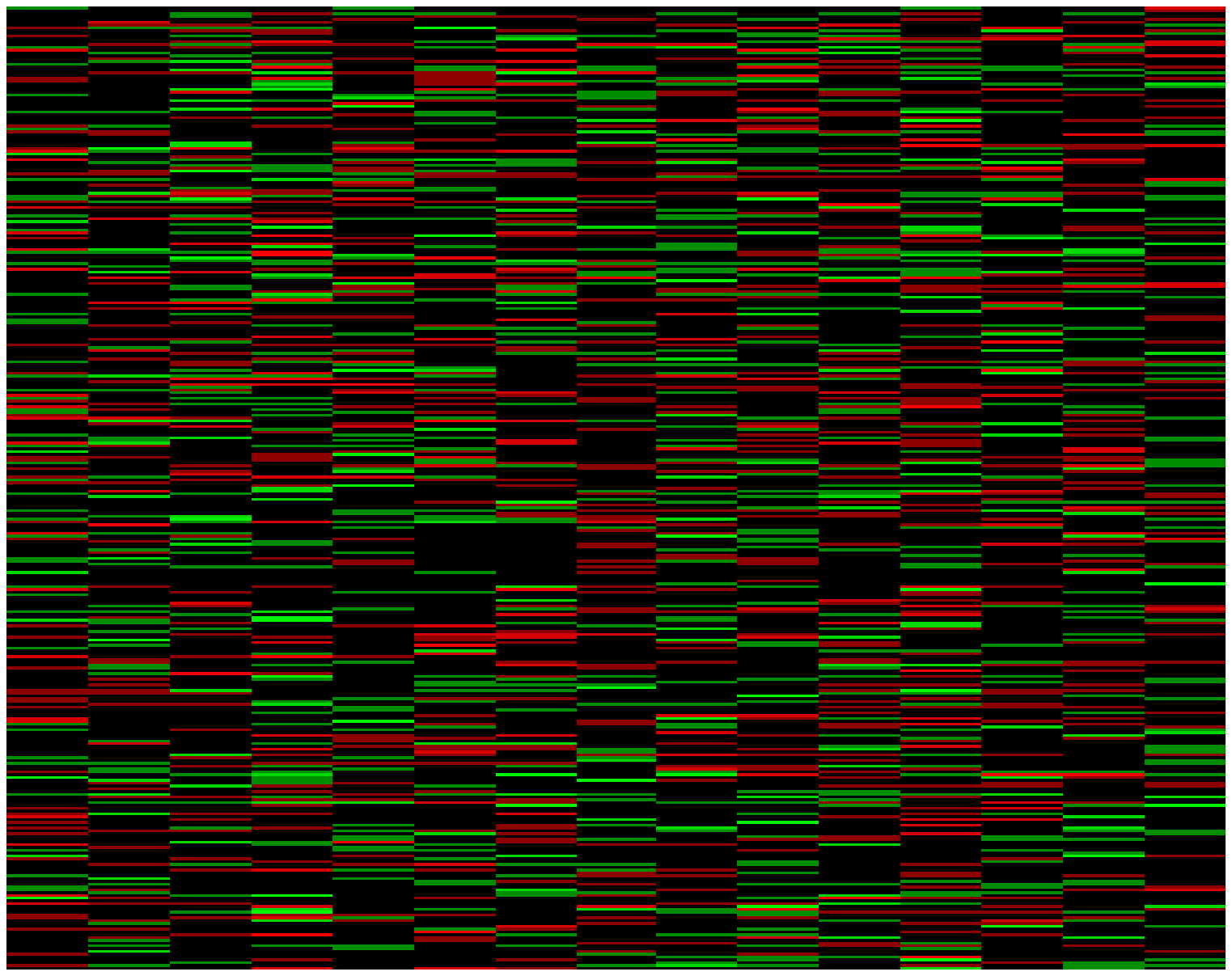}
}
\subfigure[$B$ calculated by $l_{1 \mbox{-} 2}$-GLasso]{
\includegraphics[width=6cm]{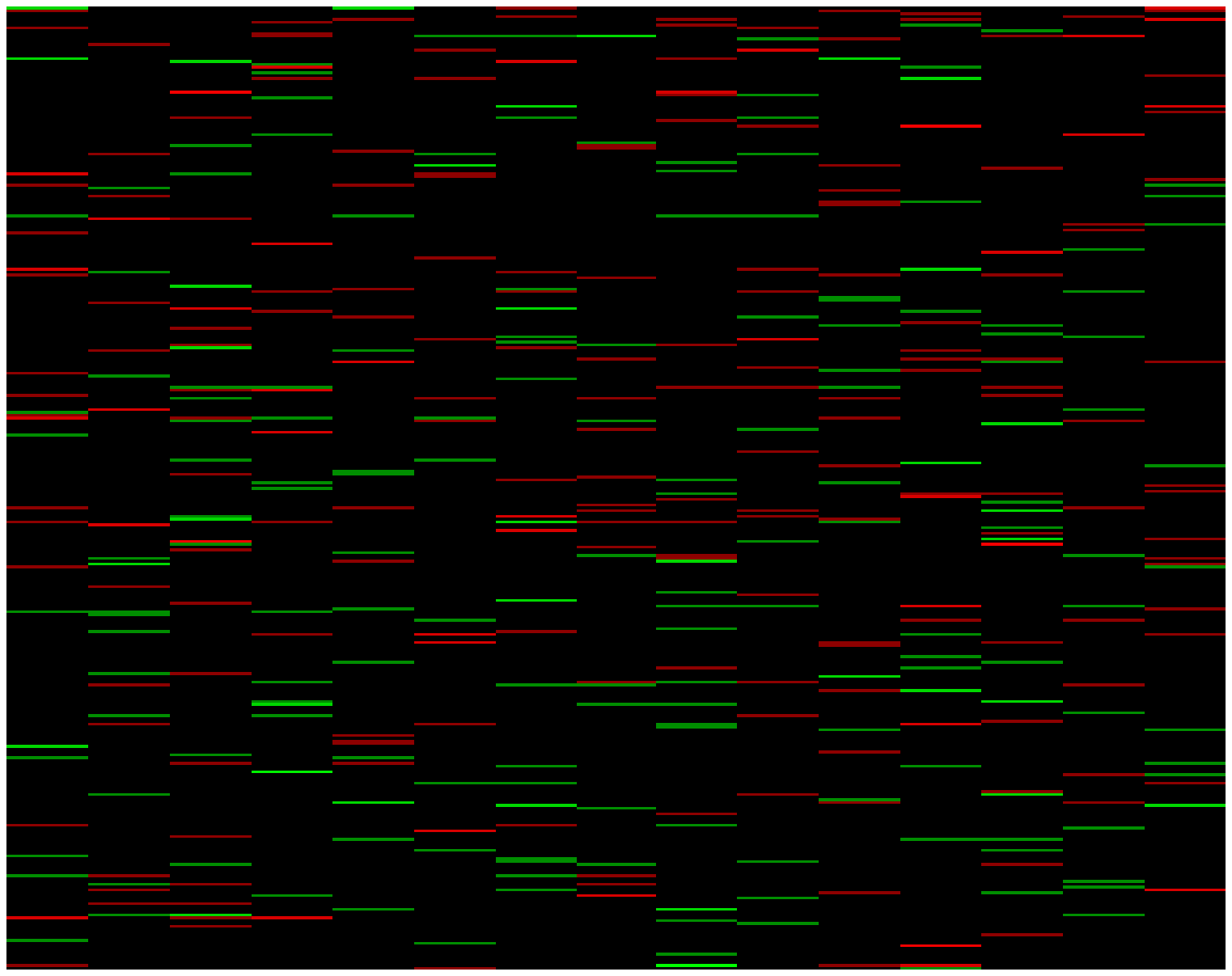}
}
\caption{We show the regression coefficient matrix calculated by ICLasso and $l_{1 \mbox{-} 2}$-GLasso. $l_{1 \mbox{-} 2}$-GLasso can get a sparser and more accurate regression coefficient matrix $B$}.
\label{f8}
\end{figure}

Back to the sparsity, we compare the regression coefficient matrix $B$ calculated by $l_{1 \mbox{-} 2}$-GLasso with ICLasso. As we can see in Figure \ref{f8}, the proposed model can get sparser solutions. For example, we set our sights on SNP rs4311. In the real dataset, the SNP rs4311 was detected to have an association with the gene InfParietal (L) \citep{wang2012identifying}. ICLasso not only finds an association between the SNP rs4311 and the gene InfParietal (L) but also detects that the SNP rs4311 is associated with gene AmygVol (L), CerebCtx (L), LatVent (L), EntCtx (L), Fusiform (L). But in our model, $l_{1 \mbox{-} 2}$-GLasso calculates the coefficients between SNP rs4311 and gene AmygVol (L), CerebCtx (L), LatVent (L), EntCtx (L), Fusiform (L) that are 0.019947, 0, 0, 0, 0. For genes that are not correlated with an SNP, most regression coefficients calculated by $l_{1 \mbox{-} 2}$-GLasso are $0$. The result shows that our model achieves sparser solutions compared with ICLasso. 

\begin{table}[H]
\begin{center}
\caption{The 24 SNPs identified in biological experiments contain 14 SNPs chosen using GFLasso}
\begin{tabular}{lcll}
\hline
\textit{SNP} & Number of Phenotype & Phenotype ID(Hemisphere) \\ \hline
\textbf{rs4311} & 1 & InfParietal (L) \\
\textbf{rs405509} & 8 & \begin{tabular}[l]{@{}l@{}}AmygVol (L), CerebWM (L), Fusiform (L), HippVol (L),\\ InfParietal (L), InfTemporal (L), MidTemporal (L), Postcentral (L) \end{tabular}\\ 
\textbf{rs666004} & 1 & InfTemporal (L) \\
\textbf{rs1433099} & 1 & CerebCtx (L) \\
\textbf{rs1473180} & 4 & CerebCtx (L) ,EntCtx (L), Fusiform (L), PostCing (L) \\
\textbf{rs1475345} & 1 & Parahipp (L) \\
\textbf{rs1568400} & 1 & Precentral (L) \\
\textbf{rs2149196} & 1 & Postcentral (L) \\
\textbf{rs2418811} & 1 & CerebWM (L) \\
\textbf{rs4935774} & 1 & CerebWM (L) \\
\textbf{rs6107516} & 1 & MidTemporal (L) \\
\textbf{rs6584307} & 1 & Parahipp (L) \\
\textbf{rs11191692} & 1 & EntCtx (L) \\
\textbf{rs12209631} & 2 & CerebCtx (L), HippVol (L) \\
\textbf{rs16924159} & 1 & PostCing (L) \\
rs1023024 & 1 & Precentral (L) \\
rs1269918 & 3 & CerebCtx (L), CerebWM (L), InfLatVent (L) \\
rs2327389 & 1 & AmygVol (L) \\
rs2418811 & 1 & CerebWM (L) \\
rs2756271 & 4 & EntCtx (L), HippVol (L), InfTemporal (L), Parahipp (L) \\
rs7219773 & 1 & Precentral (L) \\
rs9314349 & 1 & Parahipp (L) \\
rs10787010 & 7 & \begin{tabular}[l]{@{}l@{}}AmygVol (L), EntCtx (L), Fusiform (L), HippVol (L),\\ InfLatVent (L), InfTemporal (L), Precentral (L) \end{tabular}\\
rs10787011 & 1 & EntCtx (L) \\
rs11601726 & 2 & CerebWM (L), LatVent (L) \\ \hline
\label{t2}
\end{tabular}
\end{center}
\end{table}
\begin{table}[H]
\begin{center}
\caption{The 24 SNPs identified in biological experiments contain 16 SNPs chosen using MRCE}
\begin{tabular}{lcll}
\hline
\textit{SNP} & Number of Phenotype & Phenotype ID(Hemisphere) \\ \hline
 \textbf{rs4311} & 1 & InfParietal (L) \\
\textbf{rs405509} & 3 & \begin{tabular}[l]{@{}l@{}}AmygVol (L), CerebWM (L), HippVol (L) \end{tabular}\\
\textbf{rs666004} & 1 & InfTemporal (L) \\
\textbf{rs1433099} & 1 & CerebCtx (L) \\
\textbf{rs1473180} & 4 & CerebCtx (L) ,EntCtx (L), Fusiform (L), PostCing (L) \\
\textbf{rs1475345} & 1 & Parahipp (L) \\
\textbf{rs1568400} & 1 & Precentral (L) \\
\textbf{rs2149196} & 1 & Postcentral (L) \\
\textbf{rs2327389} & 1 & AmygVol (L) \\
\textbf{rs4935774} & 1 & CerebWM (L) \\
\textbf{rs6107516} & 1 & MidTemporal (L) \\
\textbf{rs6584307} & 1 & Parahipp (L) \\
\textbf{rs9314349} & 1 & Parahipp (L) \\
\textbf{rs10787010} & 2 & \begin{tabular}[l]{@{}l@{}}InfLatVent (L), Precentral (L) \end{tabular}\\
\textbf{rs11191692} & 1 & EntCtx (L) \\
\textbf{rs12209631} & 2 & CerebCtx (L), HippVol (L) \\
rs1023024 & 1 & Precentral (L) \\
rs1269918 & 3 & CerebCtx (L), CerebWM (L), InfLatVent (L) \\
rs2418811 & 1 & CerebWM (L) \\
rs2756271 & 4 & EntCtx (L), HippVol (L), InfTemporal (L), Parahipp (L) \\
rs7219773 & 1 & Precentral (L) \\
rs10787011 & 1 & EntCtx (L) \\
rs11601726 & 2 & CerebWM (L), LatVent (L) \\ 
rs16924159 & 1 & PostCing (L) \\ \hline
\label{t3}
\end{tabular}
\end{center}
\end{table}
\begin{table}[H]
\begin{center}
\caption{The 24 SNPs identified in biological experiments contain 17 SNPs chosen using ICLasso}
\begin{tabular}{lcll}
\hline
\textit{SNP} & Number of Phenotype & Phenotype ID(Hemisphere) \\ \hline
\textbf{rs4311} & 1 & InfParietal (L) \\
\textbf{rs405509} & 8 & \begin{tabular}[l]{@{}l@{}}AmygVol (L), CerebWM (L), Fusiform (L), HippVol (L),\\ InfParietal (L), InfTemporal (L), MidTemporal (L), Postcentral (L) \end{tabular}\\
\textbf{rs1433099} & 1 & CerebCtx (L) \\
\textbf{rs1473180} & 4 & CerebCtx (L) ,EntCtx (L), Fusiform (L), PostCing (L) \\
\textbf{rs1475345} & 1 & Parahipp (L) \\
\textbf{rs1568400} & 1 & Precentral (L) \\
\textbf{rs2149196} & 1 & Postcentral (L) \\
\textbf{rs2418811} & 1 & CerebWM (L) \\
\textbf{rs4935774} & 1 & CerebWM (L) \\
\textbf{rs6107516} & 1 & MidTemporal (L) \\
\textbf{rs6584307} & 1 & Parahipp (L) \\
\textbf{rs9314349} & 1 & Parahipp (L) \\
\textbf{rs10787010} & 7 & \begin{tabular}[l]{@{}l@{}}AmygVol (L), EntCtx (L), Fusiform (L), HippVol (L),\\ InfLatVent (L), InfTemporal (L), Precentral (L) \end{tabular}\\
\textbf{rs10787011} & 1 & EntCtx (L) \\
\textbf{rs11191692} & 1 & EntCtx (L) \\
\textbf{rs12209631} & 2 & CerebCtx (L), HippVol (L) \\
\textbf{rs16924159} & 1 & PostCing (L) \\ 
rs666004 & 1 & InfTemporal (L) \\
rs1023024 & 1 & Precentral (L) \\
rs1269918 & 3 & CerebCtx (L), CerebWM (L), InfLatVent (L) \\
rs2327389 & 1 & AmygVol (L) \\
rs2756271 & 4 & EntCtx (L), HippVol (L), InfTemporal (L), Parahipp (L) \\
rs7219773 & 1 & Precentral (L) \\
rs11601726 & 2 & CerebWM (L), LatVent (L) \\\hline
\label{t4}
\end{tabular}
\end{center}
\end{table}
\begin{table}[H]
\begin{center}
\caption{The 24 SNPs identified in biological experiments contain 22 SNPs chosen using $l_{1 \mbox{-} 2}$-GLasso}
\begin{tabular}{lcll}
\hline
\textit{SNP} & Number of phenotype & Phenotype ID(Hemisphere) \\ \hline
\textbf{rs4311} & 1 & InfParietal (L) \\
\textbf{rs405509} & 8 & \begin{tabular}[l]{@{}l@{}}AmygVol (L), CerebWM (L), Fusiform (L), HippVol (L),\\ InfParietal (L), InfTemporal (L), MidTemporal (L), Postcentral (L) \end{tabular}\\
\textbf{rs666004} & 1 & InfTemporal (L) \\
\textbf{rs1023024} & 1 & Precentral (L) \\
\textbf{rs1269918} & 3 & CerebCtx (L), CerebWM (L), InfLatVent (L) \\
\textbf{rs1433099} & 1 & CerebCtx (L) \\
\textbf{rs1473180} & 4 & CerebCtx (L) ,EntCtx (L), Fusiform (L), PostCing (L) \\
\textbf{rs1475345} & 1 & Parahipp (L) \\
\textbf{rs1568400} & 1 & Precentral (L) \\
\textbf{rs2149196} & 1 & Postcentral (L) \\
\textbf{rs2418811} & 1 & CerebWM (L) \\
\textbf{rs2756271} & 4 & EntCtx (L), HippVol (L), InfTemporal (L), Parahipp (L) \\
\textbf{rs4935774} & 1 & CerebWM (L) \\
\textbf{rs6107516} & 1 & MidTemporal (L) \\
\textbf{rs6584307} & 1 & Parahipp (L) \\
\textbf{rs9314349} & 1 & Parahipp (L) \\
\textbf{rs10787010} & 7 & \begin{tabular}[l]{@{}l@{}}AmygVol (L), EntCtx (L), Fusiform (L), HippVol (L),\\ InfLatVent (L), InfTemporal (L), Precentral (L) \end{tabular}\\
\textbf{rs10787011} & 1 & EntCtx (L) \\
\textbf{rs11191692} & 1 & EntCtx (L) \\
\textbf{rs11601726} & 2 & CerebWM (L), LatVent (L) \\
\textbf{rs12209631} & 2 & CerebCtx (L), HippVol (L) \\
\textbf{rs16924159} & 1 & PostCing (L) \\ 
rs2327389 & 1 & AmygVol (L) \\
rs7219773 & 1 & Precentral (L) \\\hline
\label{tab5}
\end{tabular}
\end{center}
\end{table}

\begin{table}[H]
\centering
\caption{SNPs identified by different models}
\begin{tabular}{lcccc} 
\hline
SNP        & \multicolumn{1}{c}{GFLasso} & \multicolumn{1}{c}{MRCE}   & \multicolumn{1}{c}{ICLasso} & \multicolumn{1}{c}{$l_{1 \mbox{-} 2}$-GLasso}     \\ 
\hline
rs4311     & \ding{56}  & \ding{56} & \ding{56}  & \ding{56}  \\
rs405509   & \ding{56}  & \ding{56} & \ding{56}  & \ding{56}  \\
rs666004   &                             & \ding{56} &                             & \ding{56}  \\
rs1023024  &                             &                            &                             & \ding{56}  \\
rs1269918  &                             &                            &                             & \ding{56}  \\
rs1433099  & \ding{56}  & \ding{56} & \ding{56}  & \ding{56}  \\
rs1473180  & \ding{56}  & \ding{56} & \ding{56}  & \ding{56}  \\
rs1475345  & \ding{56}  & \ding{56} & \ding{56}  & \ding{56}  \\
rs1568400  & \ding{56}  & \ding{56} & \ding{56}  & \ding{56}  \\
rs2149196  & \ding{56}  & \ding{56} & \ding{56}  & \ding{56}  \\
rs2327389  &                             & \ding{56} &                             &                             \\
rs2418811  & \ding{56}  &                            & \ding{56}  & \ding{56}  \\
rs2756271  &                             &                            &                             & \ding{56}  \\
rs4935774  &                             & \ding{56} & \ding{56}  & \ding{56}  \\
rs6107516  & \ding{56}  & \ding{56} & \ding{56}  & \ding{56}  \\
rs6584307  & \ding{56}  & \ding{56} & \ding{56}  & \ding{56}  \\
rs7219773  &                             &                            &                             &                             \\
rs9314349  &                             & \ding{56} & \ding{56}  & \ding{56}  \\
rs10787010 &                             & \ding{56} & \ding{56}  & \ding{56}  \\
rs10787011 & \ding{56}  &                            & \ding{56}  & \ding{56}  \\
rs11191692 & \ding{56}  & \ding{56} & \ding{56}  & \ding{56}  \\
rs11601726 &                             &                            &                             & \ding{56}  \\
rs12209631 & \ding{56}  & \ding{56} & \ding{56}  & \ding{56}  \\
rs16924159 & \ding{56}  &                            & \ding{56}  & \ding{56}  \\ 
\hline
Count      & \multicolumn{1}{c}{14}      & \multicolumn{1}{c}{16}     & \multicolumn{1}{c}{17}      & \multicolumn{1}{c}{22}      \\
\hline
\end{tabular}
\label{t6}
\end{table}

\section{Conclusion}
In this paper, we propose the Graphical Lasso based on difference of $l_{1}$ and $l_{2}$ norms, which introduce a new penalty for the sparsity of $B$. Based on two important assumptions, we jointly estimate the regression coefficient matrix $B$ and the output structure $\Theta$. Similar to ICLasso, the optimization problem can be solved based on existing algorithms.
Through the synthetic dataset, we demonstrate that $l_{1 \mbox{-} 2}$-GLasso outperforms other models in the recovery of the eQTL associations and the gene network structure.
The results on real datasets also show the superiority of $l_{1 \mbox{-} 2}$-GLasso and confirm the assumptions of the model. 
Also, we can use other penalty norm for $B$ and $\Theta$, like $l_{\frac{1}{2}}$ penalty which is closer to the $l_{0}$ penalty. Future work will seek higher efficiency of the solution algorithm and decomposition of large-scale problems. The application domain of this model can also be extended to financial data.

\begin{acks}[Data availability]
Data analyzed in this article are available in the R-package "bgsmtr". With the R console: 1.data(bgsmtr-example-data) 2.str(bgsmtr-example-data) 3.SNP <- t(bgsmtr-example-data-SNP-data)
4.BM <- t(bgsmtr-example-data-Brain-Measures)
\end{acks}
\bibliographystyle{imsart-nameyear} 
\bibliography{bio}       


\end{document}